%% file: acl_latex.tex
\documentclass[11pt]{article}

\usepackage[preprint]{acl}

\usepackage{times}
\usepackage{latexsym}

\usepackage[T1]{fontenc}

\usepackage[utf8]{inputenc}

\usepackage{microtype}

\usepackage{inconsolata}

\usepackage{graphicx}
\usepackage{multirow}
\usepackage{amsmath}

\usepackage{amssymb}
\usepackage{tcolorbox}
\tcbuselibrary{breakable}

\definecolor{xm-purple}{RGB}{210,210,210}
\definecolor{xm-grey}{RGB}{242,242,242}

\newtcolorbox[list inside=prompt,auto counter]{prompt}[1][]{
    colbacktitle=xm-purple!90,
    colback=xm-grey!30,
    coltitle=black,
    fontupper=\footnotesize,
    boxsep=5pt,
    left=0pt,
    right=0pt,
    top=0pt,
    bottom=0pt,
    boxrule=0.5pt,
    breakable,
    #1,
    width=\linewidth,
}

\usepackage{booktabs}
\title{\textsc{DynaRev}: Benchmarking Web Agents under Dynamic Goal Revision}

\author{
\textbf{Shiqi Zhang}$^{1,2}$,
\textbf{Guochen Zhou}$^{1}$,
\textbf{Zesheng Wei}$^{1}$,
\textbf{Hai Zhao}$^{2}$,
\textbf{Zhisong Zhang}$^{1}$ \\
$^{1}$City University of Hong Kong \quad
$^{2}$Shanghai Jiao Tong University \\
\texttt{szhang2879-c@my.cityu.edu.hk},
\texttt{zhisong.zhang@cityu.edu.hk}
}

\begin{document}
\maketitle
\begin{abstract}
Web agents are typically evaluated with goals fully specified before execution and fixed thereafter, whereas real users often revise their requirements during interaction. To study such interactions, we introduce \textsc{DynaRev}, a benchmark of 181 tasks to evaluate web agents under \emph{dynamic goal revision}. \textsc{DynaRev} uses semantic milestones to deliver revisions at task-defined stages, covers additive revisions, unconditional overrides, and runtime-conditioned overrides, and evaluates correctness, responsiveness, and efficiency through trajectory-grounded evaluation. Five-model evaluations show that agents often satisfy only part of the task requirements, with performance varying across task sources, revision types, and models. Further analyses reveal varied milestone timing, mixed effects of dynamic delivery, and contrasting revision-handling outcomes, with gains from instruction rewriting while preserving browser state and losses from browser resets. \textsc{DynaRev} provides a testbed for studying how web agents adapt to evolving user goals. The code and data are available at \url{https://github.com/zsq259/DynaRev}.

\end{abstract}

\section{Introduction}


Recent advances in large language models (LLMs) have enabled autonomous agents to perceive, reason, plan, and act in complex digital environments~\citep{10.1145/3711896.3736555,xi2025rise,mei2025aios}. Among them, web agents interact with website interfaces through human-like actions such as clicking, typing, and scrolling, allowing them to operate websites without relying on dedicated APIs~\citep{zheng2024seeact,lai2024autowebglm}. They support a wide range of practical tasks such as information seeking, online shopping, travel planning, and service navigation. With increasingly capable foundation models and agent frameworks, web agents are now able to execute complex multi-step tasks in realistic environments~\citep{ICLR2024_e91bf7df,he2024webvoyager,ICLR2025_f2c6e459}.

However, most web-agent tasks assume that the user's goal is fully specified before execution and remains fixed throughout the interaction~\citep{deng2023mind2web,zhou2024webarena,koh2024visualwebarena,he2024webvoyager,xue2025an,gou2025mind2web2}. Real-world user goals are often more dynamic. For example, a user may first ask an agent to find a flight from Los Angeles to Boston, then add ``Avoid overnight flights'' after viewing the candidates, revise this preference to ``If daytime flights cost more than \$150 extra, overnight flights are acceptable,'' and later request nonstop flights only.
Such interactions are not merely a fixed instruction split across multiple turns: later instructions may add new constraints, override previously active ones, or introduce requirements that apply conditionally based on the web state observed during execution. We refer to this setting as \emph{dynamic goal revision}.

\begin{figure}[t]
\centering
\setlength{\belowcaptionskip}{-0.6cm}
\setlength{\abovecaptionskip}{0.15cm}
\includegraphics[width=\linewidth]{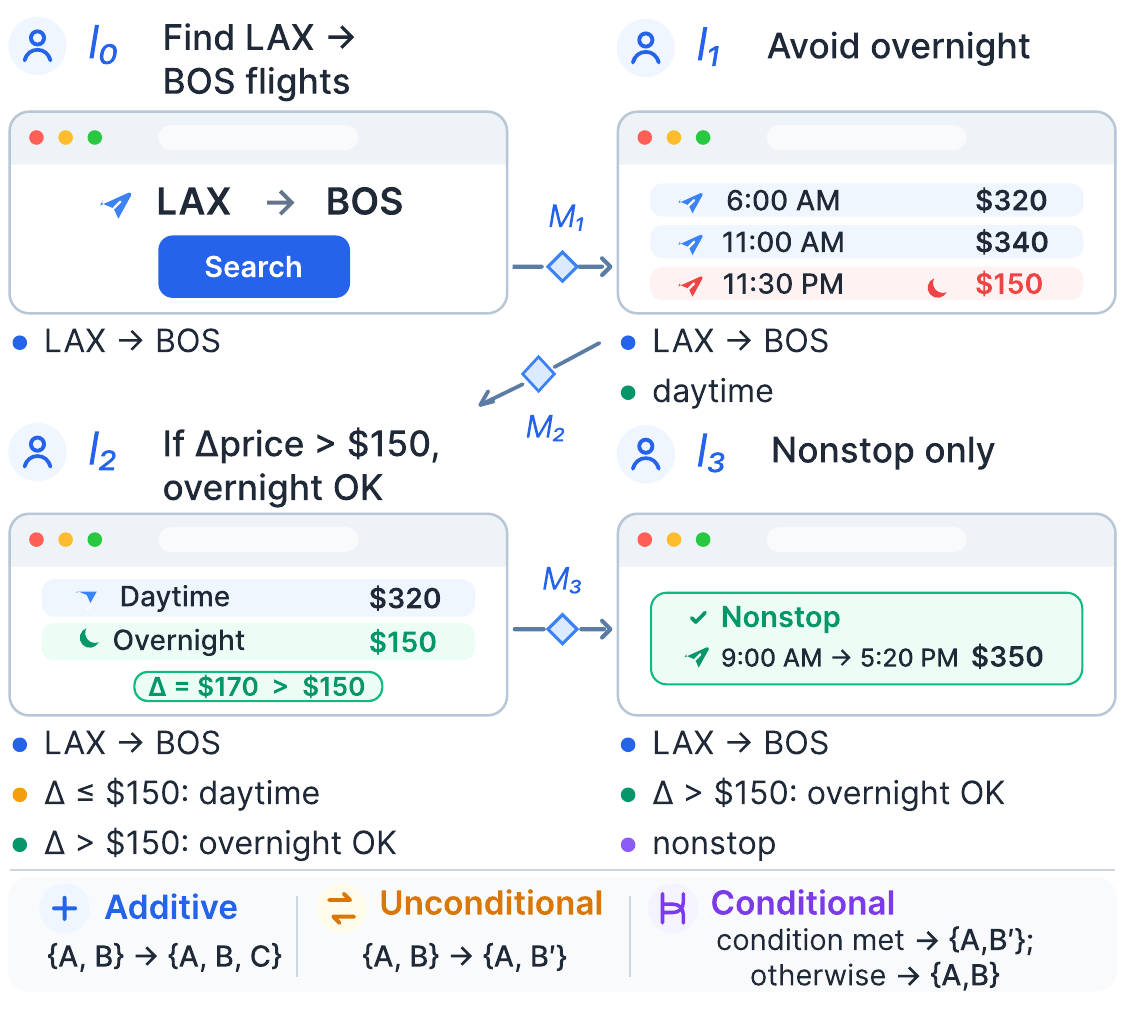}
\caption{Example dynamic goal revision task.}
\label{fig:task_overview}
\end{figure}


Recent work has begun to recognize goal adaptation as an important agent capability. AgentChangeBench~\citep{rana2025agentchangebench} studies goal shifts in conversational tool-using agents, DeskCraft~\citep{wang2026deskcraft} considers interruptions and corrections in desktop workflows, and InterruptBench~\citep{zou2026userschangemindevaluating} evaluates additions, revisions, and retractions in long-horizon web navigation.
However, three gaps remain in web settings. First, revisions are often injected based on relative trajectory progress rather than task-semantic progress, so agents following different paths may receive the same revision at different task stages. Second, existing settings largely omit conditional revisions whose applicability depends on the web state. Finally, evaluations mainly assess task-level success and recovery cost, with limited analysis of revision-level constraint satisfaction, post-revision responsiveness, and cost--quality trade-offs.


To address these gaps, we introduce \textsc{DynaRev}, a benchmark for evaluating web agents under dynamic goal revisions. DynaRev uses \emph{semantic milestones}, task-defined stages of the current goal, to trigger revisions independently of the execution path, and covers additive, unconditional-override, and conditional-override revisions, with conditional branches resolved from the web state. We develop a revision-aware, trajectory-grounded evaluation framework that assesses answer correctness, revision responsiveness, and cost-aware efficiency.


We construct 181 dynamic tasks from Online-Mind2Web~\citep{xue2025an} and Mind2Web~2~\citep{gou2025mind2web2}, and evaluate five models. Agents often satisfy only part of the task requirements, with performance varying across task sources, revision types, and models. Further analyses reveal substantial variation in milestone timing and mixed effects of dynamic delivery, while instruction rewriting with preserved browser state improves quality and efficiency and browser resets degrade performance. These results reveal behaviors missed by static evaluation.


Our contributions are threefold: (i) We formulate dynamic goal revision for web agents and introduce semantic milestone-based revision delivery. (ii) We construct \textsc{DynaRev}, covering additive revisions, unconditional overrides, and runtime-conditioned overrides. (iii) We develop a revision-aware, trajectory-grounded evaluation assessing correctness, responsiveness, and efficiency.

\begin{figure*}[t]
\centering
\setlength{\belowcaptionskip}{-0.6cm}
\setlength{\abovecaptionskip}{0.15cm}
\includegraphics[width=\linewidth]{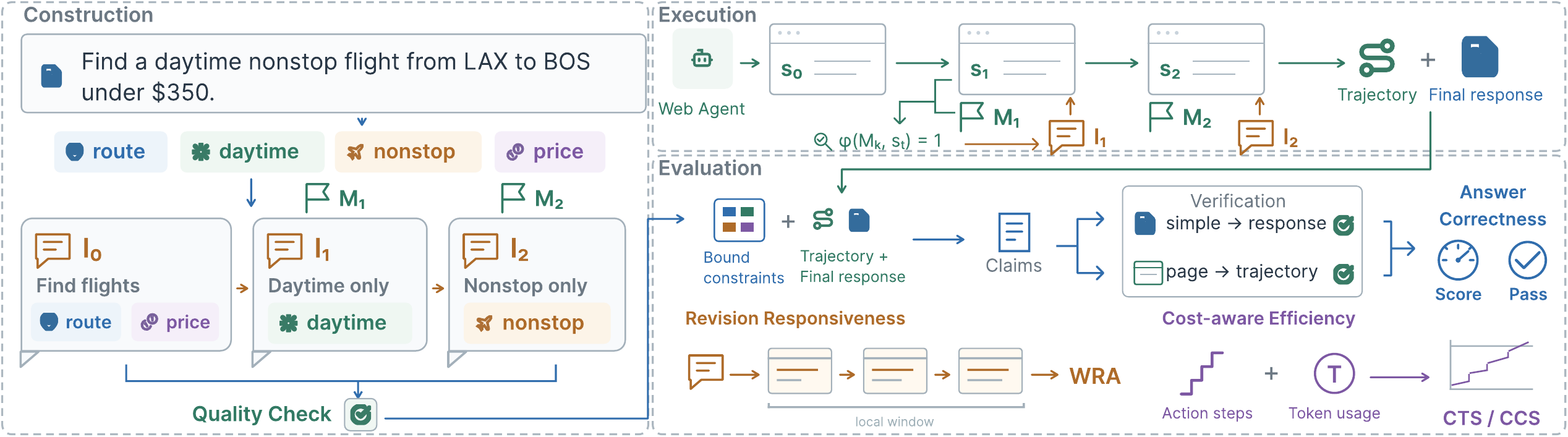}
\caption{Overview of benchmark construction, milestone-triggered execution, and revision-aware evaluation.}
\label{fig:benchmark_overview}
\end{figure*}

\section{Related Work}
\label{sec:related_work}



\paragraph{GUI Agents and Web Agents.}
GUI agents interact with graphical interfaces through actions such as clicking, typing, and scrolling to complete user tasks \citep{zhang2024large,nguyen-etal-2025-gui}. web agents instantiate this setting in browser environments, where agents can also use structured page representations such as HTML and accessibility trees. Representative systems such as SeeAct \citep{zheng2024seeact} and WebVoyager \citep{he2024webvoyager} demonstrate LLM-based agents operating on real websites. We focus on web agents as the evaluation setting in this work, while dynamic goal revision is relevant to GUI agents more broadly.

\paragraph{Web Agent Evaluation Benchmarks.}
The evaluation of web-agents has progressed from offline action prediction to interactive and real-world execution. Mind2Web~\citep{deng2023mind2web} evaluates agents against human interaction trajectories, while WebArena~\citep{zhou2024webarena} and VisualWebArena~\citep{koh2024visualwebarena} provide interactive sandbox environments with functional evaluation. 
WorkArena~\citep{workarena2024} and WorkArena++~\citep{NEURIPS2024_0b82662b} extend evaluation to enterprise workflows and compositional tasks.
WebVoyager~\citep{he2024webvoyager} extends evaluation to live websites, and more recent benchmarks such as Online-Mind2Web~\citep{xue2025an} and Mind2Web~2~\citep{gou2025mind2web2} further improve task realism and outcome verification. Despite these differences, their tasks generally specify the complete user objective before execution and keep it unchanged throughout the interaction.

\paragraph{Multi-turn Interaction and Dynamic Goal Revision.}
Multi-turn interaction has been studied in both web and general agent settings. 
MT-Mind2Web~\citep{self-map} evaluates context-dependent sequential instructions, while WebLINX~\citep{pmlr-v235-lu24e} studies real-world website navigation through multi-turn dialogue.
More recent work directly considers evolving goals: AgentChangeBench \citep{rana2025agentchangebench} studies goal shifts in conversational tool-using agents, IntentFlux~\citep{zhang2026userschangemindsmeasuring} evaluates intent drift under superseded and withdrawn requirements, while COMPASS \citep{qin2026compassbenchmarkingconstrainedoptimization} and RECAP \citep{mitra2026recap} consider evolving constraints and intent in multi-turn interactions.


Related settings have emerged for GUI agents, including ReInAgent~\citep{jia2025reinagentcontextawareguiagent}, DeskCraft~\citep{wang2026deskcraft}, and most closely, InterruptBench~\citep{zou2026userschangemindevaluating}, which evaluates user additions, revisions, and retractions during long-horizon web navigation. However, these settings do not align revision delivery with task-semantic progress or include conditional revisions whose active requirements depend on the web state. Our work addresses both with semantic milestone-based delivery and runtime-conditioned overrides.

\paragraph{Fine-grained and Process-aware Agent Evaluation.}
Recent work has moved beyond binary final-task success. Mind2Web 2 \citep{gou2025mind2web2} decomposes complex requirements into fine-grained criteria, while ProBench \citep{yang2026probench} evaluates intermediate process information. Dynamic-goal benchmarks also consider adaptation, such as goal-shift recovery in AgentChangeBench \citep{rana2025agentchangebench} and post-interruption success and execution cost in InterruptBench \citep{zou2026userschangemindevaluating}. 
Our evaluation binds atomic constraints to individual revisions and verifies them using the agent's final response and trajectory, while assessing post-revision responsiveness and cost-aware efficiency.

\section{Problem Formulation}
\label{sec:problem}

We formalize web-agent interaction under dynamic goal revisions. Throughout this section, consider a user who initially asks an agent to ``Find me a daytime flight from New York to Los Angeles,'' and may revise the request during execution.

\subsection{Web Agent Interaction}

We consider a web agent interacting with a browser environment over multiple steps. At step $t$, the agent receives observation $s_t$ and produces action $a_t$. A trajectory up to state $s_t$ is represented as

{\small
\setlength{\abovedisplayskip}{1pt}
\setlength{\belowdisplayskip}{-1pt}
\begin{equation}
\tau_t=(s_0,a_0,s_1,a_1,\ldots,s_t),
\end{equation}
}%
which records the preceding state-action sequence and ends with the latest observed state.

\subsection{Dynamic Goal Revision}

A dynamic goal revision task is defined as

{\small
\setlength{\abovedisplayskip}{1pt}
\setlength{\belowdisplayskip}{-1pt}
\begin{equation}
T =
\left(
I_0,
\{(M_k,I_k)\}_{k=1}^{n}
\right),
\end{equation}
}%
where $I_0$ is the initial instruction, each $I_k$ ($k\geq1$) is a revision delivered during execution to update the previously specified goal, and $n$ is the number of revisions. Each instruction specifies one or more task constraints, and revisions update which constraints are active.

A central challenge is deciding when a revision should be introduced. Triggering at a fixed step or at a fixed relative position along an execution trajectory can expose different agents to the same revision at substantially different stages of task progress, since successful web trajectories may vary considerably in both length and structure. 

We therefore define each milestone $M_k$ as an \emph{inevitable semantic stage} of the goal that has already been specified before revision $I_k$. Importantly, a milestone is derived from the previously specified goal rather than from the revision that will follow it. Any successful execution of the current goal should eventually reach this stage, regardless of the concrete sequence of UI actions used to get there. In the running example, a milestone can be ``The initial search for daytime flights from New York to Los Angeles has been completed and the search results have been observed.'' A later requirement such as ``Only consider nonstop flights'' is introduced after this stage is reached.

A trigger $\phi(M_k,s_t)\in\{0,1\}$ determines whether $M_k$ is reached at state $s_t$. If $\phi(M_k,s_t)=1$, revision $I_k$ is delivered before the next step.

\subsection{Revision Types}

Using the same flight request, we consider three forms of revision. An \emph{additive revision} introduces a new constraint while preserving previously active constraints, e.g., ``Only consider nonstop flights.'' An \emph{unconditional override} replaces a previously active constraint with a new one, e.g., ``Actually, choose an overnight flight instead.'' A \emph{conditional override} replaces a previously active constraint with an alternative only when a specified runtime condition is satisfied, e.g., ``If daytime flights cost more than \$150 extra, overnight flights are acceptable.'' In a conditional override, the agent must determine the applicable branch and corresponding active constraints from the web state, rather than receiving them from the revision.

\section{Benchmark Construction}
\label{sec:benchmark}

We construct dynamic goal revision tasks by transforming static goals into sequences of instructions and revisions with explicit constraint bindings, constructing semantic milestones for revision delivery, and validating task quality and interaction naturalness. Construction details are in Appendix~\ref{app:construction}.

\subsection{Static-to-Dynamic Task Transformation}

We first decompose the original task goal into atomic constraints, each describing an independently verifiable requirement. These constraints provide the basic units for revision construction and evaluation. We then construct an initial instruction $I_0$ that is sufficiently specific to form a natural and executable request, while leaving part of the original goal to be introduced during execution.

The remaining constraints are organized into a sequence of revisions according to a natural interaction order. Closely related constraints may be introduced together rather than mechanically separated into different turns. For override revisions, a previously active constraint is replaced or relaxed; conditional overrides additionally associate the replacement with a runtime condition that can be verified from the web environment. Each constraint is finally bound to the revision that introduces or modifies it, providing an explicit correspondence between goal evolution and evaluation units.

\subsection{Milestone Construction}

For each revision $I_k$, we construct its milestone $M_k$ from the cumulative goal specified by $I_0,\ldots,I_{k-1}$. The milestone describes a completed semantic stage that can be identified from the web state, while avoiding specific UI actions, reference-trajectory states, or information from the upcoming revision. We also avoid requiring a particular result configuration; for example, completing a search can constitute a milestone even when no matching candidates are returned.

\subsection{Quality Control}

Generated tasks are checked for both structural validity and interaction naturalness. The initial instruction must express a coherent executable goal, while subsequent revisions must introduce meaningful changes rather than artificial fragments of the original instruction. Milestones must describe semantic stages implied by preceding goals and remain independent of a particular reference trajectory or future revision. For conditional overrides, the triggering condition must be observable from the runtime environment. We additionally verify the consistency between revisions and their associated constraints. Tasks that fail these checks are regenerated or excluded from the benchmark.

\section{Revision-Aware Evaluation}
\label{sec:evaluation}

We evaluate agent executions through constraint verification and post-revision behavioral assessment, with metrics covering answer correctness, revision responsiveness, and cost-aware efficiency. Implementation details are in Appendix~\ref{app:evaluation}.

\subsection{Constraint Verification}

Following the Agent-as-a-Judge paradigm~\citep{zhuge2025agentasajudge,gou2025mind2web2}, we evaluate each constraint using the agent's final response and interaction trajectory. Constraints associated with undelivered revisions are counted as unsatisfied. For conditional overrides, we determine the active branch and evaluate the corresponding constraints.

Each constraint is decomposed into independently verifiable claims. According to the required evidence, we classify them into two types: \emph{simple} claims can be verified directly from the agent's final response, whereas \emph{page} claims require supporting evidence from web states in the interaction trajectory. A constraint is satisfied only if all of its claims are verified. For task $i$, we denote the resulting judgment for constraint $c$ by $z_{i,c} \in \{0,1\}$.


\subsection{Post-Revision Behavioral Assessment}

Final outcomes alone do not indicate whether an agent actually responds to a revision. For each delivered revision $I_{i,k}$, we therefore examine the subsequent local action window and use an LLM judge to determine whether the agent's behavior is consistent with the revised intent. We denote this binary judgment by $R_{i,k}\in\{0,1\}$. 

\subsection{Evaluation Metrics}

\paragraph{Answer Correctness.}
Let $\mathcal{C}_i$ be the constraints scored for task $i$. We define its constraint satisfaction score $q_i$ and task-success indicator $y_i$ as

{\small
\begin{equation}
q_i =
\frac{1}{|\mathcal{C}_i|}
\sum_{c\in\mathcal{C}_i} z_{i,c}.
\label{eq:constraint_score}
\end{equation}
}

{\small
\setlength{\belowdisplayskip}{-1pt}
\begin{equation}
y_i = \mathbf{1}[q_i=1].
\label{eq:task_success}
\end{equation}
}%
At the dataset level, we report mean constraint satisfaction and task success rate.

For revision-level analysis, let $\mathcal{C}_{i,k}$ denote the constraints introduced or modified by instruction $I_k$. We define their satisfaction score as

{\small
\setlength{\abovedisplayskip}{-2pt}
\setlength{\belowdisplayskip}{-2pt}
\begin{equation}
q_{i,k}
=
\frac{1}{|\mathcal{C}_{i,k}|}
\sum_{c\in\mathcal{C}_{i,k}} z_{i,c},
\qquad k\geq0.
\label{eq:instruction_constraint_score}
\end{equation}
}

We additionally report Mean First Failure Round (MFFR). For task $i$, let $k_i^*$ be the first revision round whose associated constraints are not all satisfied, or $n_i+1$ if no revision fails. We define

{\small
\setlength{\abovedisplayskip}{-2pt}
\setlength{\belowdisplayskip}{-2pt}
\begin{equation}
\mathrm{MFFR}=\frac{1}{N}\sum_{i=1}^{N}k_i^*.
\end{equation}
}%
Higher MFFR indicates that the agent satisfies revised requirements for more rounds.






\paragraph{Revision Responsiveness.}
Window Response Accuracy (WRA) measures whether an agent makes an appropriate behavioral adjustment immediately after a delivered revision, independently of the eventual task outcome. Let $\mathcal{R}_i^{\mathrm{fired}}$ denote the indices of revisions delivered during task $i$, and let $\mathcal{D}_{\mathrm{fired}}$ denote the set of tasks with at least one delivered revision. We first average responses within each task and then across tasks:

{\small
\setlength{\abovedisplayskip}{-2pt}
\setlength{\belowdisplayskip}{-2pt}
\begin{equation}
\mathrm{WRA}
=
\frac{1}{|\mathcal{D}_{\mathrm{fired}}|}
\sum_{i \in \mathcal{D}_{\mathrm{fired}}}
\frac{1}{|\mathcal{R}_i^{\mathrm{fired}}|}
\sum_{k \in \mathcal{R}_i^{\mathrm{fired}}}
R_{i,k}.
\end{equation}
}

\paragraph{Cost-aware Efficiency.}
We measure execution cost in two ways: the number of action steps and token usage. Let $c_i$ denote the execution cost of task $i$ under either measure. For a cost threshold $b$, we define Cumulative Task Success (CTS) and Cumulative Constraint Satisfaction (CCS) as

{\small
\setlength{\abovedisplayskip}{-2pt}
\setlength{\belowdisplayskip}{-2pt}
\begin{equation}
\mathrm{CTS}(b) =
\sum_{i=1}^{N}
y_i\,\mathbf{1}[c_i\leq b].
\label{eq:cts}
\end{equation}
}

{\small
\setlength{\abovedisplayskip}{-2pt}
\setlength{\belowdisplayskip}{-2pt}
\begin{equation}
\mathrm{CCS}(b) =
\sum_{i=1}^{N}
q_i\,\mathbf{1}[c_i\leq b].
\label{eq:ccs}
\end{equation}
}%
CTS counts fully successful tasks, while CCS additionally credits partial constraint satisfaction.

We further define cost-aware efficiency as

{\small
\setlength{\abovedisplayskip}{-2pt}
\setlength{\belowdisplayskip}{-2pt}
\begin{equation}
\mathrm{CE}
=
\frac{1}{N}
\sum_{i=1}^{N}
q_i
\left(
1-\frac{c_i}{B}
\right),
\label{eq:ce}
\end{equation}
}%
where $B$ is the maximum cost considered in the corresponding curve. CE averages correctness weighted by the remaining budget fraction after each task, jointly capturing task quality and execution efficiency. Appendix~\ref{app:ce_auc} shows its equivalence to the normalized area under the CCS curve.

\section{Experiments}
\label{sec:experiments}

\subsection{Experimental Setup}

\paragraph{Dataset.}

We construct dynamic tasks from Online-Mind2Web~\citep[OM2W;][]{xue2025an} and Mind2Web~2~\citep[M2W2;][]{gou2025mind2web2}, covering web navigation and long-horizon information synthesis.
From 70 seed tasks per source, our pipeline and manual review retains 181 tasks across three revision types, summarized in Table~\ref{tab:dataset_statistics}. Scenario details are provided in Appendix~\ref{app:dataset_details}.

\begin{table}[h]
\centering
\small
\setlength{\belowcaptionskip}{-0.6cm}
\setlength{\abovecaptionskip}{0.2cm}
\resizebox{0.49\textwidth}{!}{
\begin{tabular}{llccc}
\toprule
Source & Type & Tasks & Avg. Cons. & Avg. Rounds \\
\midrule
\multirow{3}{*}{OM2W}
& Additive & 37 & 4.76 & 2.59 \\
& Uncond.  & 39 & 4.56 & 2.92 \\
& Cond.    & 31 & 5.74 & 3.26 \\
\midrule
\multirow{3}{*}{M2W2}
& Additive & 39 & 9.69 & 2.97 \\
& Uncond.  & 23 & 10.04 & 4.22 \\
& Cond.    & 12 & 9.75 & 4.92 \\
\midrule
\multicolumn{2}{l}{Total} & 181 & 6.95 & 3.22 \\
\bottomrule
\end{tabular}
}
\caption{Dataset statistics by source and revision type. Avg. Cons. and Avg. Rounds denote average constraint and instruction-round counts. Uncond. and Cond. denote unconditional and conditional overrides.}
\label{tab:dataset_statistics}
\end{table}

\begin{table*}[t]
\centering
\setlength{\belowcaptionskip}{-0.2cm}
\small
\begin{tabular}{lccccccc}
\toprule
& \multicolumn{3}{c}{Correctness}
& \multicolumn{1}{c}{Responsiveness}
& \multicolumn{3}{c}{Efficiency} \\
\cmidrule(lr){2-4} \cmidrule(lr){5-5} \cmidrule(lr){6-8}
Model
& Score$\uparrow$
& Success$\uparrow$
& MFFR$\uparrow$
& WRA$\uparrow$
& CE (Steps)$\uparrow$
& CE (Tokens)$\uparrow$
& Avg Steps$\downarrow$ \\
\midrule
DeepSeek-V4.1-Flash
& 0.698
& \textbf{43.1\%}
& 2.21
& 66.8\%
& 0.482
& 0.637
& 43.0 \\

GPT-6.1-Sol
& 0.598
& 28.2\%
& 1.91
& 68.8\%
& 0.477
& 0.554
& \textbf{24.9} \\

Gemini-3.8-Flash
& \textbf{0.735}
& 42.5\%
& \textbf{2.28}
& \textbf{69.4\%}
& \textbf{0.522}
& \textbf{0.673}
& 33.1 \\

Qwen3.8-Flash
& 0.604
& 39.2\%
& 2.09
& 68.4\%
& 0.410
& 0.549
& 50.9 \\

Claude-Opus-5-5
& 0.681
& 36.5\%
& 2.16
& 68.3\%
& 0.448
& 0.596
& 43.8 \\
\bottomrule
\end{tabular}
\caption{Main results across all dynamic subsets.}
\label{tab:main_results}
\end{table*}


\paragraph{Baselines.}
We evaluate DeepSeek-V4.1-Flash~\citep{deepseekai2026deepseekv41flash}, GPT-6.1-Sol~\citep{openai2026gpt61sol}, Gemini-3.8-Flash~\citep{google2026gemini38flash}, Qwen3.8-Flash~\citep{qwen3.8flashnext}, and Claude-Opus-5-5~\citep{anthropic2026claudeopus55} as representative baseline web agents, abbreviated as DeepSeek, GPT, Gemini, Qwen, and Claude.


\paragraph{Implementation.}
We execute all agents with BrowserGym~\citep{chezelles2025browsergym} using viewport-restricted AXTree observations, a 100-step budget, and temperature 0. DeepSeek-V4.1-Flash is used for data construction, milestone judging, and offline evaluation. Further execution and evaluation details are in Appendices~\ref{app:execution} and~\ref{app:evaluation}.

\subsection{Main Results}

\begin{table*}[t]
\centering
\small
\setlength{\belowcaptionskip}{-0.3cm}
\resizebox{\textwidth}{!}{
\begin{tabular}{l*{12}{c}}
\toprule
& \multicolumn{6}{c}{OM2W} & \multicolumn{6}{c}{M2W2} \\
\cmidrule(lr){2-7} \cmidrule(lr){8-13}
& \multicolumn{2}{c}{Additive}
& \multicolumn{2}{c}{Unconditional}
& \multicolumn{2}{c}{Conditional}
& \multicolumn{2}{c}{Additive}
& \multicolumn{2}{c}{Unconditional}
& \multicolumn{2}{c}{Conditional} \\
\cmidrule(lr){2-3} \cmidrule(lr){4-5} \cmidrule(lr){6-7}
\cmidrule(lr){8-9} \cmidrule(lr){10-11} \cmidrule(lr){12-13}
Model
& Score$\uparrow$ & Steps$\downarrow$
& Score$\uparrow$ & Steps$\downarrow$
& Score$\uparrow$ & Steps$\downarrow$
& Score$\uparrow$ & Steps$\downarrow$
& Score$\uparrow$ & Steps$\downarrow$
& Score$\uparrow$ & Steps$\downarrow$ \\
\midrule
DeepSeek-V4.1-Flash & 0.780 & 32.4 & \textbf{0.774} & 30.6 & 0.824 & 38.5 & 0.570 & 55.4 & 0.581 & 56.5 & 0.518 & 62.2 \\
GPT-6.1-Sol & 0.793 & \textbf{19.8} & 0.642 & 27.7 & 0.753 & \textbf{23.7} & 0.404 & \textbf{27.7} & 0.435 & \textbf{23.2} & 0.401 & \textbf{29.0} \\
Gemini-3.8-Flash & 0.795 & 29.3 & 0.759 & \textbf{26.2} & \textbf{0.898} & 30.8 & \textbf{0.606} & 38.6 & \textbf{0.647} & 40.9 & \textbf{0.642} & 40.5 \\
Qwen3.8-Flash & 0.804 & 33.7 & 0.739 & 37.1 & 0.727 & 39.2 & 0.394 & 67.4 & 0.419 & 74.3 & 0.275 & 80.0 \\
Claude-Opus-5-5 & \textbf{0.858} & 30.2 & 0.697 & 38.9 & 0.853 & 31.9 & 0.480 & 58.7 & 0.568 & 59.1 & 0.516 & 54.8 \\
\bottomrule
\end{tabular}
}
\caption{Performance breakdown by task source and revision type.}
\label{tab:source_revision_breakdown}
\end{table*}

\begin{figure}[h]
\centering
\setlength{\belowcaptionskip}{-0.5cm}
\setlength{\abovecaptionskip}{-0.1cm}
\includegraphics[width=0.8\linewidth]{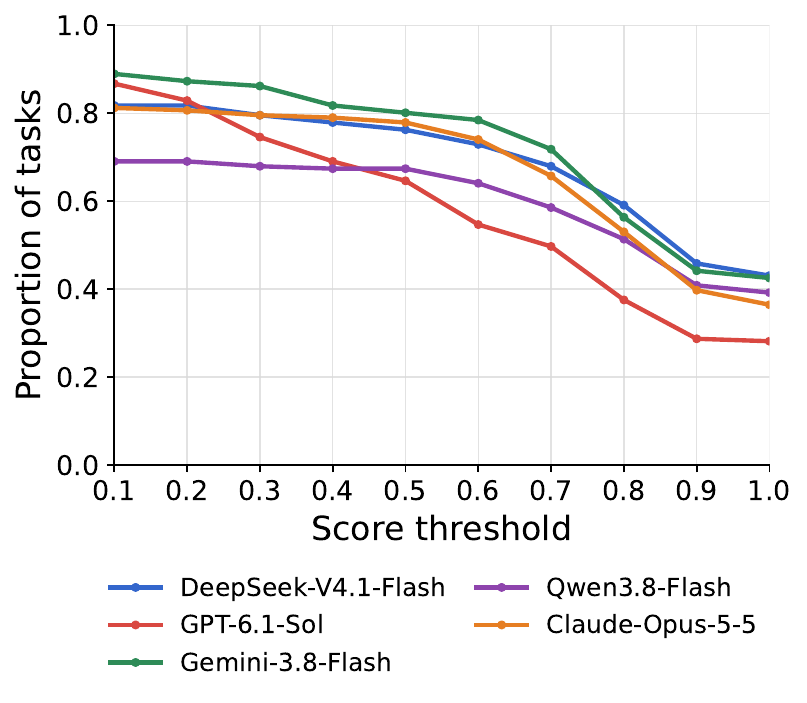}
\caption{Task-level distribution of constraint satisfaction scores. Each point reports the proportion of tasks reaching the corresponding score threshold.}
\label{fig:score_thresholds}
\end{figure}

\paragraph{Agents differ in both correctness and completion patterns.}
Table~\ref{tab:main_results} shows substantial differences in final task performance across agents. Gemini achieves the highest mean constraint satisfaction (0.735), while DeepSeek has the highest task success rate (43.1\%). However, mean satisfaction and strict completion do not induce the same ranking: Qwen has a lower Score than Claude (0.604 vs.\ 0.681), but a higher success rate (39.2\% vs.\ 36.5\%). MFFR provides a complementary sequential view: Gemini maintains revised requirements for the most rounds before the first failure (2.28), whereas GPT fails earliest on average (1.91). Figure~\ref{fig:score_thresholds} shows that Claude performs better at low and intermediate thresholds, while Qwen retains more tasks at the highest thresholds.

\begin{figure*}[htbp]
\centering
\setlength{\belowcaptionskip}{-0.3cm}
\setlength{\abovecaptionskip}{0.05cm}
\includegraphics[width=1\linewidth]{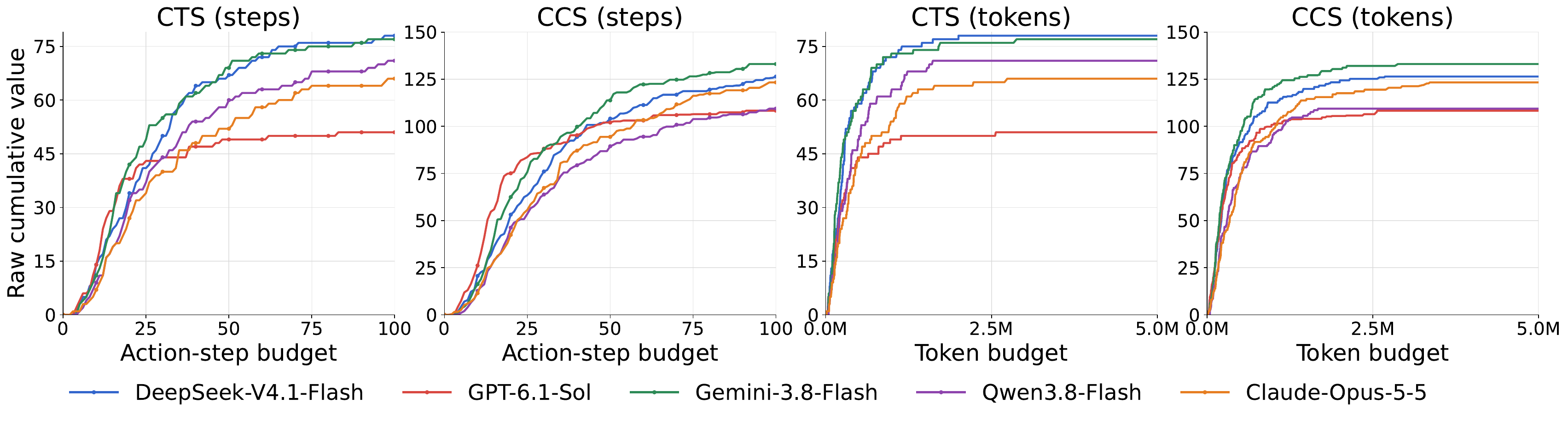}
\caption{Cost--quality curves under action-step and token budgets.}
\label{fig:cost_curves}
\end{figure*}

\paragraph{Performance degrades with task complexity, while revision-type effects are context dependent.}
Table~\ref{tab:source_revision_breakdown} breaks down performance by task source and revision type. M2W2 yields lower scores for every model and revision type and generally requires more action steps, consistent with its higher constraint counts and longer revision sequences. Revision-type effects, however, interact strongly with task source. 
Conditional tasks are not the weakest OM2W subset for four of the five models; on M2W2, they produce the lowest score for three of the five agents, including 0.275 for Qwen. Overall, task complexity produces a consistent performance gap, whereas differences among revision types vary across sources and agents.

\begin{table}[h]
\centering
\setlength{\abovecaptionskip}{0.3cm}
\setlength{\belowcaptionskip}{-0.6cm}
\small
\begin{tabular}{lcc}
\toprule
& \multicolumn{2}{c}{Revision Satisfaction $q_{i,k}$} \\
\cmidrule(lr){2-3}
Model
& WRA $=1$
& WRA $=0$ \\
\midrule
DeepSeek-V4.1-Flash
& 0.683
& 0.644 \\
GPT-6.1-Sol
& 0.587
& 0.465 \\
Gemini-3.8-Flash
& 0.720
& 0.696 \\
Qwen3.8-Flash
& 0.609
& 0.577 \\
Claude-Opus-5-5
& 0.677
& 0.628 \\
\bottomrule
\end{tabular}
\caption{Revision-level analysis conditioned on WRA.}
\label{tab:wra_conditioned_satisfaction}
\end{table}

\paragraph{Immediate responses do not consistently translate into satisfied revisions.}
WRA measures whether an agent makes a locally appropriate behavioral adjustment immediately after a revision, independently of the eventual task outcome. Although WRA ranges from 66.8\% to 69.4\% across models, its ordering does not closely follow final correctness. Table~\ref{tab:wra_conditioned_satisfaction} shows that revisions with WRA${}=1$ have higher final satisfaction for all five agents, but the gap is modest for most models, and many responsive revisions remain unsatisfied. This divergence shows that immediate behavioral adaptation and eventual revision satisfaction capture different aspects of performance.

\paragraph{Models exhibit distinct cost--quality trade-offs.}
Accounting for execution cost reveals different budget--performance profiles across agents. GPT uses the fewest action steps (24.9) but has the lowest overall Score (0.598), whereas DeepSeek achieves higher correctness (0.698) with more steps (43.0); nevertheless, their step-based CE is nearly identical (0.477 vs.\ 0.482). Gemini combines the highest correctness with the second-lowest average action count (33.1 steps) and therefore achieves the highest CE under both action-step and token costs. The cost--quality curves in Figure~\ref{fig:cost_curves} further show that GPT accumulates performance rapidly under small budgets but plateaus relatively early, while Gemini and DeepSeek continue to improve as additional budget becomes available.

\section{Analysis}
\label{sec:analysis}

\subsection{Static vs.\ Dynamic Delivery}

We compare 30 additive tasks with their original single-turn counterparts, with 15 tasks from each source. The two settings share the same final requirements and execution budget, differing only in whether the requirements are given upfront or progressively during execution. Table~\ref{tab:static_dynamic} shows no uniform effect on correctness: GPT and Qwen improve on both sources, Claude declines on both, while Gemini shows opposite trends across the two sources. Across all model--task pairs, 32.0\% improve, 28.0\% worsen, and 40.0\% remain unchanged. Despite mixed correctness effects, dynamic delivery typically requires more actions.

\begin{table}[h]
\centering
\small
\setlength{\belowcaptionskip}{-0.5cm}
\begin{tabular}{@{}lrrrr@{}}
\toprule
& \multicolumn{2}{c}{OM2W}
& \multicolumn{2}{c}{M2W2} \\
\cmidrule(lr){2-3}
\cmidrule(lr){4-5}
Model
& $\Delta$Score & $\Delta$Steps
& $\Delta$Score & $\Delta$Steps \\
\midrule
DeepSeek & $-0.025$ & $+9.06$  & $-0.035$ & $+6.47$ \\
GPT      & $+0.079$ & $-0.74$  & $+0.122$ & $+4.87$ \\
Gemini   & $-0.133$ & $+12.54$ & $+0.108$ & $+7.53$ \\
Qwen     & $+0.161$ & $+8.54$  & $+0.115$ & $-12.67$ \\
Claude   & $-0.107$ & $+14.34$ & $-0.169$ & $+10.93$ \\
\bottomrule
\end{tabular}
\caption{Changes from static to dynamic additive delivery on paired tasks. $\Delta$ denotes dynamic minus static.}
\label{tab:static_dynamic}
\end{table}

Trajectory inspection reveals two competing effects. Progressive delivery can reach the final goal through intermediate stages, but it can also introduce rework, interface loops, and budget exhaustion. Thus, dynamic delivery is neither uniformly beneficial nor harmful; instead, it exposes model-dependent execution behaviors that static evaluation may not reveal. Full results are in Appendix~\ref{app:static_dynamic}.

\begin{figure*}[t]
\centering
\setlength{\belowcaptionskip}{-0.3cm}
\includegraphics[width=1\linewidth]{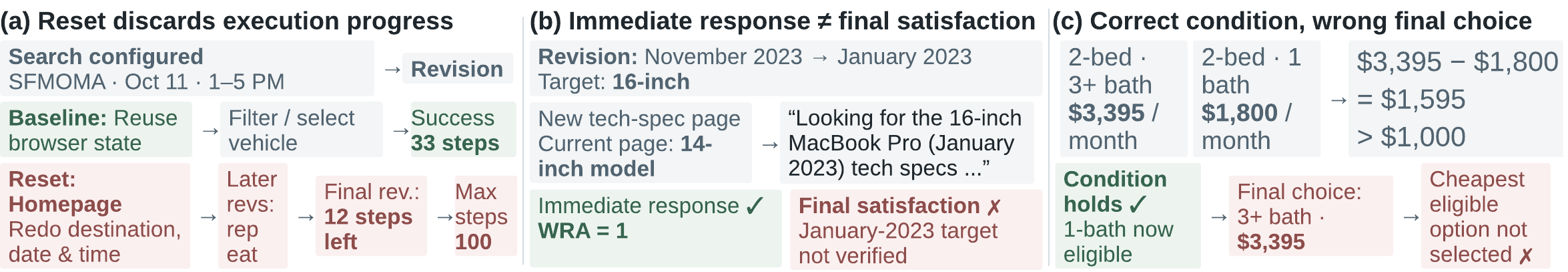}
\caption{Representative failure cases under dynamic goal revision.}
\label{fig:failure_cases}
\end{figure*}

\subsection{Revision Handling Strategies}
\label{sec:revision_handling}


When a revision is delivered, a natural alternative to continuing from the current execution is to reformulate the updated goal and restart the task. We therefore compare revision-handling strategies along three dimensions: \emph{instruction representation} (original vs.\ rewritten), \emph{interaction history} (retained vs.\ cleared), and \emph{browser state} (continued vs.\ reset). We consider six meaningful combinations, excluding configurations that reset the browser while retaining interaction history. The baseline retains the original instructions, interaction history, and browser state, corresponding to direct continuation from the current execution. We evaluate all strategies with DeepSeek-V4.1-Flash on the same subset of benchmark tasks. Strategy changes are applied only when a revision is delivered, and all configurations share the same 100-step budget. Table~\ref{tab:revision_handling_strategies} reports task quality and execution cost.

\begin{table}[h]
\centering
\setlength{\belowcaptionskip}{-0.3cm}
\small
\resizebox{\linewidth}{!}{
\begin{tabular}{lccc}
\toprule
Strategy
& Score$\uparrow$
& Steps$\downarrow$
& Tokens (k)$\downarrow$ \\
\midrule
Baseline
& 0.647 & 50.3 & 806.7 \\
Rewrite
& \textbf{0.718} & \textbf{44.9} & \textbf{658.8} \\
Clear
& 0.648 & 53.6 & 667.2 \\
Clear + Rewrite
& 0.663 & 52.8 & 673.4 \\
Reset
& 0.586 & 59.7 & 691.7 \\
Reset + Rewrite
& 0.525 & 61.5 & 843.6 \\
\bottomrule
\end{tabular}
}
\caption{Comparison of revision handling strategies. Rewrite consolidates instructions, Clear removes interaction history, and Reset resets the browser.}
\label{tab:revision_handling_strategies}
\end{table}

Rewriting the accumulated instructions while preserving the ongoing execution gives the best overall trade-off, increasing Score from 0.647 to 0.718 while reducing both action steps (50.3 to 44.9) and token usage (806.7k to 658.8k). Clearing interaction history alone leaves task quality nearly unchanged (0.648 vs.\ 0.647) and reduces token usage, but requires more actions. In contrast, resetting the browser consistently worsens the overall quality--cost trade-off: Reset lowers Score to 0.586 and increases steps to 59.7, while Reset + Rewrite further reduces Score to 0.525 and requires the most steps. Comparing Clear with Reset, and Clear + Rewrite with Reset + Rewrite, indicates that retaining the current browser state provides substantial reusable execution progress. Overall, the results favor reorganizing the accumulated goal while continuing from the current execution, rather than treating each revision as a new task.

\subsection{Revision Injection Timing}

A simpler alternative to semantic milestone triggering is to inject revisions at a fixed relative position along the execution trajectory. We therefore examine where milestone-based revision timing falls along relative trajectory progress on a subset of tasks. We focus on the first revision. For task $i$ and agent $a$, let $t_{i,a}$ denote the revision-injection step and $L_{i,a}$ the completion length of an execution with only the initial instruction $I_0$. We define

{\small
\setlength{\abovedisplayskip}{-2pt}
\setlength{\belowdisplayskip}{-2pt}
\begin{equation}
r_{i,a}=\min\left(\frac{t_{i,a}}{L_{i,a}},1\right),
\end{equation}
}%
with values above 1 clipped to 1.

\begin{figure}[h]
\centering
\setlength{\belowcaptionskip}{-0.3cm}
\setlength{\abovecaptionskip}{0.1cm}
\includegraphics[width=1\linewidth]{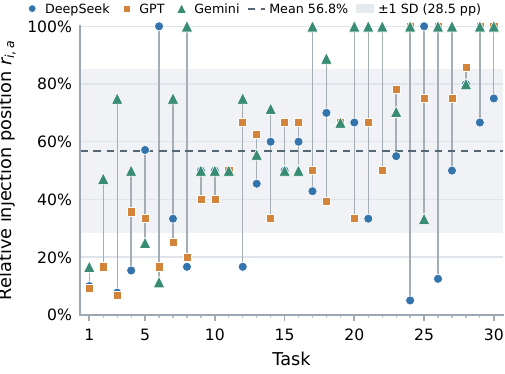}
\caption{Relative revision-injection positions across tasks and agents, ordered by mean across agents.}
\label{fig:injection_positions}
\end{figure}

Figure~\ref{fig:injection_positions} shows substantial variation in relative injection positions across tasks, with pronounced cross-agent differences for many tasks. The mean is 56.8\%, but the standard deviation reaches 28.5 percentage points, indicating that the average position masks a broad distribution of milestone timing. Thus, even though the mean lies near 60\%, a fixed trajectory percentage does not closely represent the same task stage across individual executions.

We further quantify cross-agent variation by computing $\max_a r_{i,a}-\min_a r_{i,a}$ for each task. The median range is 36.4 and mean 41.3; 70.0\% of tasks exceed 20, and 46.7\% exceed 40. The same fixed percentage may therefore deliver a revision when different agents have reached substantially different task stages, making cross-agent comparisons less controlled. Semantic milestones instead anchor revision delivery to a task-defined stage. We also performed a manual inspection of the milestone-trigger timing, with detailed results in Appendix~\ref{app:milestone_validation}.

\subsection{Failure Mode Case Studies}

We inspect trajectories to understand the quantitative patterns. Figure~\ref{fig:failure_cases} shows three representative failure modes. First, browser resets can discard execution progress, forcing repeated work until the step budget is exhausted. Second, immediate adaptation does not ensure eventual satisfaction: an agent may respond appropriately to a revision yet still violate the revised constraint. Third, an agent may correctly resolve a condition but fail to act accordingly, selecting a more expensive option despite identifying a newly eligible cheaper one.

\section{Conclusion}
\label{sec:conclusion}

We introduced \textsc{DynaRev}, a benchmark to assess web agents under dynamic goal revisions, featuring semantic milestone triggers, runtime-conditioned overrides, and fine-grained evaluation of correctness, responsiveness, and efficiency. Experiments with five models show that immediate adaptation does not ensure final constraint satisfaction. Further analyses reveal timing mismatches between fixed-position and milestone-based revisions, while instruction rewriting while preserving browser state improves task quality and efficiency. Our benchmark provides a foundation for studying web agents' ability to handle evolving user goals.

\section*{Limitations}

We acknowledge the limitations of this work. (i) Our benchmark is constructed from static web tasks and may not capture the full diversity of real-world goal changes. (ii) Milestone triggering and evaluation rely on imperfect LLM-based semantic judgments. (iii) Semantic milestones assume that meaningful intermediate task stages can be identified, which may be harder for tasks with highly diverse successful trajectories.




\bibliography{custom}

\appendix

\section{Benchmark Construction Details}
\label{app:construction}

This section provides additional details for the static-to-dynamic task transformation described in Section~4. The construction procedure follows five stages: constraint decomposition, initial instruction construction, revision and milestone organization, quality control, and constraint binding.

\subsection{Constraint Decomposition}

We first decompose each source task into a set of atomic constraints. Each constraint describes a single independently verifiable requirement and is designed to avoid semantic overlap with the others. The resulting constraints serve as the shared units for subsequent revision construction and evaluation.

Tasks with too few constraints to support meaningful multi-round decomposition are excluded at this stage. This prevents the construction procedure from producing artificial revisions by fragmenting a goal that is naturally expressed in a single turn.

\subsection{Initial Instruction Construction}

The atomic constraints are separated according to whether they are necessary for forming a natural initial request. We refer to constraints that are needed to specify a coherent and executable starting goal as \emph{required} constraints, while constraints that can be naturally introduced later are treated as \emph{later} constraints.

The initial instruction $I_0$ contains the \emph{required} constraints. If these constraints alone leave the target underspecified, a small number of \emph{later} constraints may also be included to make the request sufficiently clear. The objective is to keep $I_0$ executable without prematurely exposing requirements that can naturally arise during subsequent interaction.

Before constructing the initial instruction, we also check whether the target website is accessible. Tasks affected by persistent access failures, abnormal redirections, or similar issues that prevent meaningful execution are excluded from further construction.

\subsection{Revision and Milestone Organization}

\paragraph{Revision organization.}
The remaining constraints are organized into a sequence of revisions according to a natural interaction order. Closely related constraints may be introduced together when they would normally be expressed in the same user turn, rather than being mechanically separated to increase the number of revisions. Additive revisions directly introduce previously unspecified requirements while preserving the constraints that are already active.

\paragraph{Milestone construction.}
Each revision $I_k$ is paired with a semantic milestone $M_k$ that determines when it is delivered. The milestone is derived only from the cumulative goal specified by $I_0,\ldots,I_{k-1}$, rather than from the upcoming revision $I_k$. We describe milestones as \emph{completed semantic stages} of the currently specified task rather than particular UI actions or reference-trajectory states.

The milestone description also avoids unnecessarily requiring a specific result configuration. For example, completing a search stage can constitute a milestone even if the search returns no matching candidates. This distinction prevents milestone triggering from depending on incidental properties of the current web state rather than on task progress itself.

\paragraph{Eligible override constraints.}
For tasks containing override revisions, we distinguish between \emph{core} and \emph{preference} constraints. \emph{Core} constraints define the fundamental task object or objective, such as the target product category, search subject, or destination. \emph{Preference} constraints describe attributes that users may naturally revise during execution, such as color, price range, rating, or ranking preference. Override revisions are constructed only from \emph{preference} constraints.

\paragraph{Unconditional overrides.}
The construction of an unconditional override is separated into replacement planning and timeline organization. We first select an eligible preference constraint and choose an alternative value from the same semantic dimension. Replacement candidates are primarily collected from values observed in the source tasks; when the available candidates are insufficient, additional alternatives in the same semantic dimension are generated and reviewed for consistency with the task context.

The selected original preference and its replacement are then incorporated into the multi-round timeline. The original preference is introduced before the override, while the later revision explicitly replaces it with the new requirement. Separating replacement selection from timeline generation reduces interference between deciding \emph{what} should change and deciding \emph{when and how} that change should be expressed.

\paragraph{Conditional overrides.}
Conditional overrides retain the original preference as the default requirement and activate an alternative only when a specified runtime condition is satisfied. The condition must be observable from the web environment during execution and should introduce an additional environment-state judgment rather than reducing to a direct restatement of the preference replacement itself.

We consider two types of conditions. \emph{Numerical-threshold conditions} are defined over observable quantities such as price differences, rating differences, or distance, and activate the alternative branch when the relevant value crosses a specified threshold. \emph{Existence conditions} depend on whether a particular type of option is available, such as whether an item is in stock or whether a direct flight exists.

For conditional-override construction, the selected replacement pair is combined with a condition appropriate to the task context and constraint dimension, and the resulting branch structure is then incorporated into the multi-round timeline.

\subsection{Quality Control}

Generated tasks are reviewed for both structural validity and interaction naturalness. The initial instruction must specify a coherent executable goal, while subsequent revisions must introduce meaningful changes rather than artificial fragments of the source instruction. Milestones must describe semantic stages implied by previously specified goals, remain independent of the upcoming revision, and avoid dependence on a particular sequence of UI operations.

For conditional overrides, the runtime condition must be observable from the web environment and the alternative branch must remain meaningful in the underlying task context. We also check the consistency of the revision sequence and the constraints that it is intended to introduce or modify.

When an instance fails these checks, the relevant construction stage is regenerated. Tasks that remain invalid after the prescribed retries are excluded from the final benchmark.

\subsection{Constraint Binding}

After a task passes quality control, each atomic constraint is bound to the instruction that introduces or modifies it. This correspondence records the provenance of each evaluation requirement and allows the evaluation pipeline to distinguish constraints associated with different stages of goal evolution.

Constraint binding is performed as a separate post-processing stage rather than jointly with revision generation. Generating revisions and bindings simultaneously may encourage artificially regular partitions in which each revision is constructed merely to correspond to a convenient subset of constraints. Separating the two stages allows revision construction to prioritize natural interaction structure while preserving an explicit instruction--constraint correspondence for later evaluation.

\section{Execution Details}
\label{app:execution}

\subsection{Interaction Context}

At each step, the agent receives the accumulated user instructions, the complete action history, and web observations from the interaction trajectory. To control context growth, we retain the full web observations for only the five most recent interaction steps. Earlier observations are replaced with a fixed placeholder, while their corresponding actions remain in the context. Thus, the complete instruction and action histories are preserved throughout execution, while detailed web-state history is maintained within a fixed recent window.

\subsection{Revision Delivery and Termination}

Revisions are delivered when their corresponding semantic milestones are reached. If the agent declares completion while revisions remain pending and execution budget remains, the completion is intercepted, the next pending revision is delivered, and execution continues.

A run terminates when the agent calls \texttt{done} after all revisions have been delivered, or when the maximum execution step limit is reached.

\subsection{Environment Configuration}

All agents are executed through BrowserGym with Playwright-controlled headless Chromium. Observations use the viewport-restricted AXTree representation. The same execution configuration, including the context-management strategy, is applied to all evaluated agents.

\section{Evaluation Details}
\label{app:evaluation}

This section provides implementation details omitted from Section~\ref{sec:evaluation}, focusing on claim construction, trajectory evidence retrieval, and post-revision behavioral assessment.

\subsection{Claim Construction and Verification}

For each constraint to be evaluated, the claim generator receives the constraint, the agent's final response, and the task context, and produces one or more independently verifiable claims together with their verification types. For constraints containing relative-time expressions, we additionally provide the task execution date to resolve the temporal reference.

Claim decomposition is used when a single constraint requires multiple verification instances. For example, if a constraint requires every returned hotel to provide free breakfast, the evaluator generates a separate claim for each returned hotel. This allows each hotel to be verified independently while preserving the original constraint as a single task requirement.

Claims of type \emph{simple} are verified directly from the agent's final response, whereas \emph{page} claims are passed to the trajectory evidence retrieval procedure described below.

\subsection{Trajectory Evidence Retrieval}

For each page claim, we first identify the subject whose property or web evidence must be verified. We then inspect the final recorded web state. If it already provides sufficient evidence, the claim is verified directly from this state.

Otherwise, we search the trajectory beginning from the delivery of the instruction to which the constraint is bound. The claim subject is used as an anchor for lightweight lexical matching over subsequent web states, producing a small set of candidate states. Only these candidates are passed to the page verifier.

A lexical match is used only for candidate retrieval and is not itself treated as evidence. Candidate states are examined in trajectory order, and the first state judged to provide sufficient evidence is used to verify the claim. If no retrieved state provides sufficient evidence, the page claim is treated as unverified.

\subsection{Post-Revision Behavioral Assessment}

For each delivered revision, we examine at most the next three agent actions, stopping earlier if another revision is delivered or the task terminates. The resulting action window is evaluated by the behavioral judge to determine whether the agent responds consistently with the revised intent.

\section{Equivalence between Cost-aware Efficiency and CCS AUC}
\label{app:ce_auc}

For the CCS curve defined in Eq.~\ref{eq:ccs},

{\small
\begin{equation}
\begin{aligned}
\frac{1}{NB}\int_0^B \mathrm{CCS}(b)\,db
&=
\frac{1}{NB}
\int_0^B
\sum_{i=1}^{N}
q_i\,\mathbf{1}[c_i\leq b]\,db \\
&=
\frac{1}{NB}
\sum_{i=1}^{N}
q_i
\int_0^B
\mathbf{1}[c_i\leq b]\,db \\
&=
\frac{1}{NB}
\sum_{i=1}^{N}
q_i(B-c_i) \\
&=
\frac{1}{N}
\sum_{i=1}^{N}
q_i
\left(
1-\frac{c_i}{B}
\right) \\
&=
\mathrm{CE}.
\end{aligned}
\label{eq:ce_auc}
\end{equation}
}

The last expression is exactly the cost-aware efficiency score defined in Eq.~\ref{eq:ce}. Therefore, CE is equivalent to the normalized area under the CCS curve.

\section{Dataset Details}
\label{app:dataset_details}

\begin{figure}[h]
\centering
\includegraphics[width=1\linewidth]{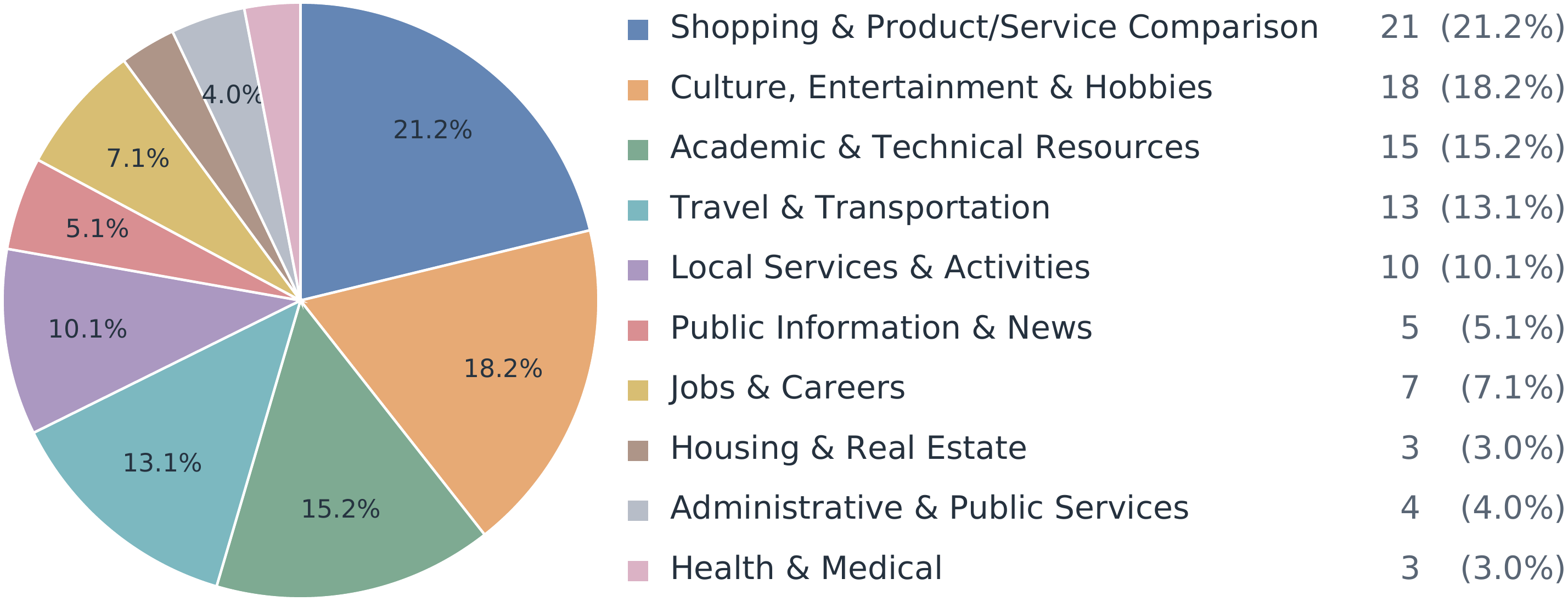}
\caption{Scenario distribution of retained seed tasks.}
\label{fig:scenario_distribution}
\end{figure}

Among the 140 sampled seed tasks, 99 contribute at least one retained dynamic task to the final benchmark. We assign each retained seed task to one scenario category and report their proportions in Figure~\ref{fig:scenario_distribution}.

\section{Static vs.\ Dynamic Delivery}
\label{app:static_dynamic}

\begin{table}[h]
\centering
\small
\setlength{\tabcolsep}{4pt}
\begin{tabular}{lrrr}
\toprule
Model & Better & Worse & Unchanged \\
\midrule
DeepSeek & 8  & 9  & 13 \\
GPT      & 15 & 9  & 6  \\
Gemini   & 9  & 7  & 14 \\
Qwen     & 11 & 3  & 16 \\
Claude   & 5  & 14 & 11 \\
\midrule
All      & 48 & 42 & 60 \\
\bottomrule
\end{tabular}
\caption{Pairwise Score changes under dynamic delivery.}
\label{tab:static_dynamic_pairs}
\end{table}

\begin{table*}[t]
\centering
\small
\begin{tabular}{lrrrrrrrrrr}
\toprule
Model
& $\geq$0.1
& $\geq$0.2
& $\geq$0.3
& $\geq$0.4
& $\geq$0.5
& $\geq$0.6
& $\geq$0.7
& $\geq$0.8
& $\geq$0.9
& $\geq$1.0 \\
\midrule
DeepSeek-V4.1-Flash
& 81.8\%
& 81.8\%
& 79.6\%
& 77.9\%
& 76.2\%
& 72.9\%
& 68.0\%
& \textbf{59.1\%}
& \textbf{45.9\%}
& \textbf{43.1\%} \\
GPT-6.1-Sol
& 86.7\%
& 82.9\%
& 74.6\%
& 69.1\%
& 64.6\%
& 54.7\%
& 49.7\%
& 37.6\%
& 28.7\%
& 28.2\% \\
Gemini-3.8-Flash
& \textbf{89.0\%}
& \textbf{87.3\%}
& \textbf{86.2\%}
& \textbf{81.8\%}
& \textbf{80.1\%}
& \textbf{78.5\%}
& \textbf{71.8\%}
& 56.4\%
& 44.2\%
& 42.5\% \\
Qwen3.8-Flash
& 69.1\%
& 69.1\%
& 68.0\%
& 67.4\%
& 67.4\%
& 64.1\%
& 58.6\%
& 51.4\%
& 40.9\%
& 39.2\% \\
Claude-Opus-5-5
& 81.2\%
& 80.7\%
& 79.6\%
& 79.0\%
& 77.9\%
& 74.0\%
& 65.7\%
& 53.0\%
& 39.8\%
& 36.5\% \\
\bottomrule
\end{tabular}
\caption{Task-level distribution of constraint satisfaction scores. Each column reports the percentage of tasks reaching the corresponding score threshold.}
\label{tab:score_thresholds}
\end{table*}

We report the full results for the static--dynamic comparison in Table~\ref{tab:static_dynamic_full}. The experiment uses the same 30 paired additive tasks as in Section~\ref{sec:analysis}, with 15 tasks from each source. Static denotes the original single-turn task, while Dynamic denotes its additive counterpart with requirements delivered progressively.

\begin{table*}[t]
\centering
\small
\begin{tabular}{lcccccccc}
\toprule
& \multicolumn{4}{c}{OM2W}
& \multicolumn{4}{c}{M2W2} \\
\cmidrule(lr){2-5}
\cmidrule(lr){6-9}
& \multicolumn{2}{c}{Score}
& \multicolumn{2}{c}{Steps}
& \multicolumn{2}{c}{Score}
& \multicolumn{2}{c}{Steps} \\
\cmidrule(lr){2-3}
\cmidrule(lr){4-5}
\cmidrule(lr){6-7}
\cmidrule(lr){8-9}
Model
& Static & Dynamic
& Static & Dynamic
& Static & Dynamic
& Static & Dynamic \\
\midrule
DeepSeek
& 0.909 & 0.884 & 20.67 & 29.73
& 0.616 & 0.581 & 45.00 & 51.47 \\
GPT
& 0.791 & 0.870 & 18.07 & 17.33
& 0.307 & 0.429 & 26.20 & 31.07 \\
Gemini
& 0.953 & 0.820 & 16.13 & 28.67
& 0.534 & 0.642 & 30.47 & 38.00 \\
Qwen
& 0.822 & 0.983 & 25.13 & 33.67
& 0.337 & 0.452 & 74.80 & 62.13 \\
Claude
& 0.939 & 0.832 & 20.13 & 34.47
& 0.738 & 0.570 & 43.00 & 53.93 \\
\bottomrule
\end{tabular}
\caption{Full results for static and dynamic additive delivery on paired tasks.}
\label{tab:static_dynamic_full}
\end{table*}

\section{Score Threshold Results}
\label{app:score_thresholds}

Table~\ref{tab:score_thresholds} reports the detailed values underlying Figure~\ref{fig:score_thresholds}.

\section{Round-wise Results}
\label{app:roundwise_results}

For diagnostic analysis, we additionally group constraint satisfaction and behavioral responsiveness by instruction round. Let $\mathcal{D}_k$ denote the tasks for which $q_{i,k}$ is defined. We compute

{\small
\begin{equation}
\mathrm{CSS}_k
=
\frac{1}{|\mathcal{D}_k|}
\sum_{i\in\mathcal{D}_k} q_{i,k}.
\end{equation}
}

For $k\geq1$, $\mathrm{CSS}_k$ measures satisfaction of constraints introduced or modified by the $k$-th revision, while $\mathrm{CSS}_0$ corresponds to the initial instruction.

Let $\mathcal{D}_k^{\mathrm{fired}}$ denote the tasks in which the $k$-th revision is delivered. We similarly define

{\small
\begin{equation}
\mathrm{WRA}_k
=
\frac{1}{|\mathcal{D}_k^{\mathrm{fired}}|}
\sum_{i\in\mathcal{D}_k^{\mathrm{fired}}}
R_{i,k}.
\end{equation}
}

Table~\ref{tab:roundwise_response} reports both quantities by round. Since later rounds contain fewer observations, we pool rounds 4 and above into $I_{4+}$.

\section{Milestone Trigger Validation}
\label{app:milestone_validation}

We conduct a small manual inspection of milestone-trigger timing on 30 sampled cases. For each trigger, annotators inspect the milestone description, the web state at which the revision is triggered, and the immediately preceding state. They independently answer two binary questions: (Q1) whether the milestone is satisfied at the trigger state, and (Q2) whether it was already satisfied at the preceding state. We derive a three-way timing label from these judgments: a trigger is \emph{early} if the milestone is not yet satisfied at the trigger state, \emph{aligned} if it is satisfied at the trigger state but not at the preceding state, and \emph{late} if it was already satisfied one step earlier.

\begin{table}[h]
\centering
\small
\begin{tabular}{lrrr}
\toprule
Annotator & Early & Aligned & Late \\
\midrule
Annotator 1 & 0.0\% & 90.0\% & 10.0\% \\
Annotator 2 & 10.0\% & 76.7\% & 13.3\% \\
\bottomrule
\end{tabular}
\caption{Human judgments of milestone-trigger timing.}
\label{tab:milestone_human_labels}
\end{table}

Both annotators judge the large majority of sampled triggers as aligned, although they differ on several cases near the semantic transition boundary. Table~\ref{tab:milestone_agreement} reports their agreement. They agree on 25 of 30 three-way timing labels (83.3\%), with Cohen's $\kappa=0.438$. Agreement is higher on the two underlying binary judgments, reaching 90.0\% for both Q1 and Q2.

\begin{table}[h]
\centering
\small
\begin{tabular}{lrr}
\toprule
Judgment & Agreement & Cohen's $\kappa$ \\
\midrule
Q1: Trigger state & 90.0\% & 0.000 \\
Q2: Previous state & 90.0\% & 0.516 \\
Three-way label & 83.3\% & 0.438 \\
\bottomrule
\end{tabular}
\caption{Agreement between the two human annotators.}
\label{tab:milestone_agreement}
\end{table}

\begin{table*}[h]
\centering
\small
\begin{tabular}{llrrrrr}
\toprule
Model & Metric & $I_0$ & $I_1$ & $I_2$ & $I_3$ & $I_{4+}$ \\
\midrule
\multirow{2}{*}{DeepSeek-V4.1-Flash}
& CSS & 70.8\% & 68.2\% & 62.9\% & 59.3\% & 50.3\% \\
& WRA & -- & 64.8\% & 68.9\% & 61.3\% & \textbf{72.0\%} \\
\midrule
\multirow{2}{*}{GPT-6.1-Sol}
& CSS & 63.1\% & 60.6\% & 50.0\% & 45.5\% & 28.5\% \\
& WRA & -- & 69.1\% & 63.9\% & 50.0\% & 46.7\% \\
\midrule
\multirow{2}{*}{Gemini-3.8-Flash}
& CSS & \textbf{74.3\%} & \textbf{76.2\%} & \textbf{64.2\%} & 57.0\% & \textbf{68.7\%} \\
& WRA & -- & 69.9\% & 70.3\% & 61.3\% & 65.6\% \\
\midrule
\multirow{2}{*}{Qwen3.8-Flash}
& CSS & 61.8\% & 58.0\% & 55.8\% & 50.4\% & 37.6\% \\
& WRA & -- & 67.8\% & \textbf{70.7\%} & \textbf{65.5\%} & 71.4\% \\
\midrule
\multirow{2}{*}{Claude-Opus-5-5}
& CSS & 70.8\% & 67.1\% & 60.9\% & \textbf{63.3\%} & 40.2\% \\
& WRA & -- & \textbf{70.6\%} & 63.2\% & 59.3\% & 69.2\% \\
\bottomrule
\end{tabular}
\caption{Round-wise constraint satisfaction score (CSS) and window response accuracy (WRA).}
\label{tab:roundwise_response}
\end{table*}

\input{appendix_prompts}


\end{document}

%% file: appendix_prompts.tex
\section{Prompt Templates}
\label{app:prompts}

\subsection{Task Construction}

\subsubsection{Constraint Extraction}
\begin{prompt}
\raggedright
\sloppy
\setlength{\emergencystretch}{3em}


\textbf{System}\par\smallskip
{\ttfamily
\noindent You are an expert at analyzing task instructions. Your job is to extract the constraints that an evaluator can actually check.\par
}\medskip


\textbf{User}\par\smallskip
{\ttfamily
\noindent Given the following task instruction, extract all constraints and output a JSON object.\par
\smallskip
\noindent A constraint is a condition, requirement, or filter stated in the instruction that can be checked from the final answer and/or a cited page.\par
\smallskip
\noindent Output format:\par
\noindent ```json\par
\noindent \{\{\par
\noindent \hspace*{1.0em}"original\_instruction": "...",\par
\noindent \hspace*{1.0em}"constraints": [\par
\noindent \hspace*{2.0em}\{\{"id": 1, "description": "..."\}\},\par
\noindent \hspace*{2.0em}\{\{"id": 2, "description": "..."\}\}\par
\noindent \hspace*{1.0em}]\par
\noindent \}\}\par
\noindent ```\par
\smallskip
\noindent Rules:\par
\noindent - Each constraint must describe exactly one thing \textemdash{} do not merge multiple constraints into one\par
\noindent - Write each constraint as a clear, self-contained statement\par
\noindent - Preserve logical structure inside a single constraint when the original instruction expresses alternatives or conjunctions.\par
\noindent - If the original instruction contains an OR relation (e.g., "A or B"), keep it as ONE composite constraint with the OR relation explicit.\par
\noindent - Do not split one OR condition into two separate atomic constraints.\par
\noindent - Extract what the instruction asks for, at the same scope it was written. Do not invent extra requirements or promote a vague overall goal into a per-item rule.\par
\noindent - Skip wording that is only subjective or scene-setting (e.g. fun, nice, good). If a sentence mixes that with a concrete requirement, keep the concrete part.\par
\smallskip
\noindent Here are some examples:\par
\smallskip
\noindent **Example 1:**\par
\noindent Task instruction: `Show me the list of Men's Blazers, Black, Size M on Uniqlo.`\par
\noindent ```json\par
\noindent \{\{\par
\noindent \hspace*{1.0em}"original\_instruction": "Show me the list of Men's Blazers, Black, Size M on Uniqlo.",\par
\noindent \hspace*{1.0em}"constraints": [\par
\noindent \hspace*{2.0em}\{\{"id": 1, "description": "Product type must be Blazers"\}\},\par
\noindent \hspace*{2.0em}\{\{"id": 2, "description": "Must be from the Men's category"\}\},\par
\noindent \hspace*{2.0em}\{\{"id": 3, "description": "Color must be Black"\}\},\par
\noindent \hspace*{2.0em}\{\{"id": 4, "description": "Size must be M"\}\}\par
\noindent \hspace*{1.0em}]\par
\noindent \}\}\par
\noindent ```\par
\smallskip
\noindent **Example 2:**\par
\noindent Task instruction: `Find Mac external hard drives with capacity greater than 2TB, sorted by price from lowest to high, showing availability.`\par
\noindent ```json\par
\noindent \{\{\par
\noindent \hspace*{1.0em}"original\_instruction": "Find Mac external hard drives with capacity greater than 2TB, sorted by price from lowest to high, showing availability.",\par
\noindent \hspace*{1.0em}"constraints": [\par
\noindent \hspace*{2.0em}\{\{"id": 1, "description": "Product type must be external hard drives"\}\},\par
\noindent \hspace*{2.0em}\{\{"id": 2, "description": "Must be compatible with Mac"\}\},\par
\noindent \hspace*{2.0em}\{\{"id": 3, "description": "Capacity must be greater than 2TB"\}\},\par
\noindent \hspace*{2.0em}\{\{"id": 4, "description": "Results must be sorted by price from lowest to highest"\}\},\par
\noindent \hspace*{2.0em}\{\{"id": 5, "description": "Availability must be shown"\}\}\par
\noindent \hspace*{1.0em}]\par
\noindent \}\}\par
\noindent ```\par
\smallskip
\noindent **Example 3:**\par
\noindent Task instruction: `I am a DC Comics fan. Could you find three comic books from different series written by the same author as Injustice: Gods Among Us Year One that feature Batman (excluding Injustice itself)? Please provide the links to the homepage of Injustice and these comic books on DC comic's official website.`\par
\noindent ```json\par
\noindent \{\{\par
\noindent \hspace*{1.0em}"original\_instruction": "I am a DC Comics fan. Could you find three comic books from different series written by the same author as Injustice: Gods Among Us Year One that feature Batman (excluding Injustice itself)? Please provide the links to the homepage of Injustice and these comic books on DC comic's official website.",\par
\noindent \hspace*{1.0em}"constraints": [\par
\noindent \hspace*{2.0em}\{\{"id": 1, "description": "Comics must be found on DC.com"\}\},\par
\noindent \hspace*{2.0em}\{\{"id": 2, "description": "The comics must be written by the same author as Injustice: Gods Among Us Year One"\}\},\par
\noindent \hspace*{2.0em}\{\{"id": 3, "description": "Each of the comics kept after Year One must feature Batman"\}\},\par
\noindent \hspace*{2.0em}\{\{"id": 4, "description": "The final answer must identify exactly three comic books besides Year One"\}\},\par
\noindent \hspace*{2.0em}\{\{"id": 5, "description": "The three comic books must be from different series"\}\},\par
\noindent \hspace*{2.0em}\{\{"id": 6, "description": "Injustice: Gods Among Us Year One itself must be excluded from the three comic books"\}\},\par
\noindent \hspace*{2.0em}\{\{"id": 7, "description": "The final answer must include a DC.com page link for Injustice: Gods Among Us Year One"\}\},\par
\noindent \hspace*{2.0em}\{\{"id": 8, "description": "The final answer must include a DC.com page link for each of the three comic books"\}\}\par
\noindent \hspace*{1.0em}]\par
\noindent \}\}\par
\noindent ```\par
\smallskip
\noindent Now extract constraints from the following task instruction:\par
\smallskip
\noindent Task instruction: \{task\}\par
}\medskip

\end{prompt}

\subsubsection{Initial Instruction Construction}
\begin{prompt}
\raggedright
\sloppy
\setlength{\emergencystretch}{3em}


\textbf{System}\par\smallskip
{\ttfamily
\noindent You are an expert at analyzing website homepages and determining what information is required to initiate a search or action.\par
}\medskip


\textbf{User}\par\smallskip
{\ttfamily
\noindent You are given a list of task constraints and a description of a website's homepage.\par
\smallskip
\noindent Step 1: For each constraint, classify whether it is REQUIRED by the homepage to initiate the first search or action, or whether it can be added LATER after seeing results.\par
\smallskip
\noindent A constraint is REQUIRED only if the homepage has a form field or input that MUST be filled with this information before the user can proceed. Preferences, filters, and secondary refinements that do not block initial navigation are LATER.\par
\smallskip
\noindent Step 2: Design I0 as the minimum but naturally complete starting request.\par
\smallskip
\noindent **Principle:**\par
\noindent - I0 should be phrased as a natural user instruction that expresses a complete intent\par
\noindent - Start with REQUIRED constraints; if they alone form a natural complete request, use only those\par
\noindent - If REQUIRED constraints alone would result in an unnatural or vague request that obscures the user's core intent,   you may supplement by selecting constraints from LATER that clarify the main goal or subject matter\par
\noindent - Avoid over-supplementing; only add what is necessary to express the user's genuine intent\par
\noindent - Do NOT make I0 over-complete by including every constraint. If a constraint can naturally be refined after initial results   (e.g., source/site preference, ranking preference, secondary filters), prefer leaving it for later turns.\par
\noindent - I0 must explicitly include the core topic/subject of the task; avoid generic openings that only describe an action\par
\noindent - Avoid underspecified I0 such as "search for products", "look for guides", or other action-only phrasing without the task's key subject\par
\smallskip
\noindent The key criterion is whether I0 would sound like something a real user would naturally say when starting the task, with their core intent clearly expressed.\par
\smallskip
\noindent \{previous\_step4\_context\}\par
\smallskip
\noindent Homepage description:\par
\noindent \{homepage\_description\}\par
\smallskip
\noindent Constraints:\par
\noindent \{constraints\}\par
\smallskip
\noindent Output format:\par
\noindent ```json\par
\noindent \{\{\par
\noindent \hspace*{1.0em}"classification": [\par
\noindent \hspace*{2.0em}\{\{"id": 1, "required": true, "reason": "homepage has a destination field"\}\},\par
\noindent \hspace*{2.0em}\{\{"id": 2, "required": false, "reason": "this is a filter, not a homepage field"\}\}\par
\noindent \hspace*{1.0em}],\par
\noindent \hspace*{1.0em}"i0": "..."\par
\noindent \}\}\par
\noindent ```\par
\smallskip
\noindent Here are some examples:\par
\smallskip
\noindent **Example 1:**\par
\noindent Homepage description: Trip.com train search form (Trains tab). Departure station and arrival station inputs must be filled before Search will run. The form also has a departure date/time control, but it already has a default. High-speed vs other train types is not a required homepage field.\par
\smallskip
\noindent Constraints:\par
\noindent ```json\par
\noindent \{\{\par
\noindent \hspace*{1.0em}"constraints": [\par
\noindent \hspace*{2.0em}\{\{"id": 1, "description": "Must be a high speed train ticket"\}\},\par
\noindent \hspace*{2.0em}\{\{"id": 2, "description": "Departure time must be before 23:00"\}\},\par
\noindent \hspace*{2.0em}\{\{"id": 3, "description": "Origin must be Shanghai"\}\},\par
\noindent \hspace*{2.0em}\{\{"id": 4, "description": "Destination must be Beijing"\}\}\par
\noindent \hspace*{1.0em}]\par
\noindent \}\}\par
\noindent ```\par
\smallskip
\noindent ```json\par
\noindent \{\{\par
\noindent \hspace*{1.0em}"classification": [\par
\noindent \hspace*{2.0em}\{\{"id": 1, "required": false, "reason": "high-speed is not a blocking field on the train search form"\}\},\par
\noindent \hspace*{2.0em}\{\{"id": 2, "required": false, "reason": "departure date/time already has a default; before 23:00 is not required to run Search"\}\},\par
\noindent \hspace*{2.0em}\{\{"id": 3, "required": true, "reason": "departure-station field must be filled before Search can run"\}\},\par
\noindent \hspace*{2.0em}\{\{"id": 4, "required": true, "reason": "arrival-station field must be filled before Search can run"\}\}\par
\noindent \hspace*{1.0em}],\par
\noindent \hspace*{1.0em}"i0": "Find a high speed train ticket from Shanghai to Beijing."\par
\noindent \}\}\par
\noindent ```\par
\smallskip
\noindent **Example 2:**\par
\noindent Homepage description: DC.com comics catalog. Titles can be browsed without filling the search box. Character, series, quantity, and link details are later refinements.\par
\smallskip
\noindent Constraints:\par
\noindent ```json\par
\noindent \{\{\par
\noindent \hspace*{1.0em}"constraints": [\par
\noindent \hspace*{2.0em}\{\{"id": 1, "description": "Comics must be found on DC.com"\}\},\par
\noindent \hspace*{2.0em}\{\{"id": 2, "description": "The comics must be written by the same author as Injustice: Gods Among Us Year One"\}\},\par
\noindent \hspace*{2.0em}\{\{"id": 3, "description": "Each of the comics kept after Year One must feature Batman"\}\},\par
\noindent \hspace*{2.0em}\{\{"id": 4, "description": "The final answer must identify exactly three comic books besides Year One"\}\},\par
\noindent \hspace*{2.0em}\{\{"id": 5, "description": "The three comic books must be from different series"\}\},\par
\noindent \hspace*{2.0em}\{\{"id": 6, "description": "Injustice: Gods Among Us Year One itself must be excluded from the three comic books"\}\},\par
\noindent \hspace*{2.0em}\{\{"id": 7, "description": "The final answer must include a DC.com page link for Injustice: Gods Among Us Year One"\}\},\par
\noindent \hspace*{2.0em}\{\{"id": 8, "description": "The final answer must include a DC.com page link for each of the three comic books"\}\}\par
\noindent \hspace*{1.0em}]\par
\noindent \}\}\par
\noindent ```\par
\smallskip
\noindent ```json\par
\noindent \{\{\par
\noindent \hspace*{1.0em}"classification": [\par
\noindent \hspace*{2.0em}\{\{"id": 1, "required": false, "reason": "already on DC.com; not a blocking homepage field"\}\},\par
\noindent \hspace*{2.0em}\{\{"id": 2, "required": false, "reason": "catalog can be browsed without filling search; author is not a required homepage field"\}\},\par
\noindent \hspace*{2.0em}\{\{"id": 3, "required": false, "reason": "character filter after seeing the author's comics"\}\},\par
\noindent \hspace*{2.0em}\{\{"id": 4, "required": false, "reason": "quantity can be decided after results"\}\},\par
\noindent \hspace*{2.0em}\{\{"id": 5, "required": false, "reason": "series diversity is a later refinement"\}\},\par
\noindent \hspace*{2.0em}\{\{"id": 6, "required": false, "reason": "exclusion is a later refinement"\}\},\par
\noindent \hspace*{2.0em}\{\{"id": 7, "required": false, "reason": "answer-format link, not needed to start the search"\}\},\par
\noindent \hspace*{2.0em}\{\{"id": 8, "required": false, "reason": "answer-format links, not needed to start the search"\}\}\par
\noindent \hspace*{1.0em}],\par
\noindent \hspace*{1.0em}"i0": "On DC.com, find Injustice: Gods Among Us Year One and other comic books by the same author."\par
\noindent \}\}\par
\noindent ```\par
\smallskip
\noindent Now classify and design I0.\par
}\medskip

\end{prompt}

\subsubsection{Additive Revision and Milestone Construction}
\begin{prompt}
\raggedright
\sloppy
\setlength{\emergencystretch}{3em}


\textbf{System}\par\smallskip
{\ttfamily
\noindent You are an expert at designing natural multi-turn user interactions for GUI tasks. Your job is to arrange a list of constraints into a natural multi-turn instruction sequence, simulating how a real user would progressively refine their request during task execution. For each subsequent instruction, you also define a milestone that determines when it should be injected.\par
}\medskip


\textbf{User}\par\smallskip
{\ttfamily
\noindent Given the original task instruction, its extracted constraints, and the pre-designed I0, complete the multi-turn instruction sequence and define milestones.\par
\smallskip
\noindent \#\# Part 1: Instruction Sequence\par
\smallskip
\noindent **I0:**\par
\noindent - Already provided \textemdash{} do not modify it\par
\smallskip
\noindent **Subsequent instructions:**\par
\noindent - Introduce remaining constraints (those NOT in I0) in the order a real user would naturally think of them after seeing intermediate results\par
\noindent - A constraint should be introduced later if the user would only think of it after seeing some results or making progress\par
\noindent - Do NOT repeat or restate constraints that are already clearly expressed in I0\par
\noindent - Each turn after I0 must introduce one or more remaining Step1 constraints not already covered by earlier turns. Do not add requirements beyond the provided constraints, including extra filters, optimization goals, or output details.\par
\noindent - If I0 already covers all constraints, do not invent a follow-up; return only I0.\par
\noindent - Avoid weak increments such as tone strengthening, exclusion of obvious negatives, or restating the same constraint with different wording \textemdash{} these do not count as new revisions\par
\noindent - Group constraints that a user would naturally say together in the same turn. Do NOT split constraints of the same type across multiple turns just to inflate the count\par
\noindent - Each Step1 constraint should be introduced exactly once by one instruction. I0 covers some constraints; subsequent instructions should cover the rest.\par
\noindent - For composite constraints (including OR/AND relations), treat the full logic expression as one unit and introduce it in one instruction without changing its operators.\par
\noindent - Do not rewrite logical type across turns (OR must remain OR; AND must remain AND).\par
\noindent - Avoid artificial decomposition: if only one natural follow-up remains, keep the sequence concise but still natural.\par
\smallskip
\noindent **Style:**\par
\noindent - All instructions should be written in natural user voice, as if the user is speaking directly\par
\noindent - Use phrasing like "Show me...", "Only show...", "Now set..." rather than task description style\par
\smallskip
\noindent \#\# Part 2: Milestones\par
\smallskip
\noindent For each instruction i\_k (k >= 1), define a milestone that triggers its injection.\par
\smallskip
\noindent A milestone is a **branch-robust semantic task state** reached after correctly completing i0 through i\_\{\{k-1\}\}. It should describe user-visible task progress, not UI mechanics or system internals.\par
\smallskip
\noindent How to produce each milestone:\par
\noindent 1. Summarize what prior instructions have semantically established (intent/state, not actions).\par
\noindent 2. Phrase that as a user-understandable task state that remains valid across common branches   (including empty-result branches).\par
\noindent 3. Check two conditions: (a) saying i\_k at this state would not feel abrupt for the user, and (b) among stages satisfying (a), this is an earlier one that is reliably reachable by diverse agent trajectories.\par
\smallskip
\noindent Quality guidance:\par
\noindent - Derive the milestone only from i0..i\_\{\{k-1\}\}; never leak content from i\_k.\par
\noindent - Prefer semantic state wording over UI-surface wording.\par
\noindent - Avoid implementation-specific language such as "system processed/registered/executed...".\par
\noindent - A milestone should hold across all branches that correct prior execution can produce. When a concrete phrasing like "X is displayed" is branch-robust for this task, use it; otherwise phrase more general.\par
\noindent - Necessity is a baseline, not a goal \textemdash{} multiple stages on a correct completion path usually satisfy it. Prefer an earlier candidate over one that restates i0's full terminal condition, since the latter collapses the milestone into an i0-completion check.\par
\smallskip
\noindent Examples illustrating the principle:\par
\noindent - "The search has been constrained to the previously requested criteria, including cases with zero matches" \textemdash{} rather than "A non-empty result list is displayed".\par
\noindent - "The query scope defined by prior constraints is established" \textemdash{} rather than "The system has registered the user's criteria".\par
\noindent - "The sofa browsing context is active with black leather candidates visible among current results" \textemdash{} rather than "A filtered list of black leather sofas with three seats is displayed".\par
\noindent - "Eligibility criteria for child benefit are displayed" \textemdash{} on a static info page where the content always renders, the concrete phrasing is itself branch-robust.\par
\smallskip
\noindent Output format:\par
\noindent ```json\par
\noindent \{\{\par
\noindent \hspace*{1.0em}"instructions": [\par
\noindent \hspace*{2.0em}\{\{"id": 0, "instruction": "..."\}\},\par
\noindent \hspace*{2.0em}\{\{"id": 1, "instruction": "...", "milestone": "..."\}\},\par
\noindent \hspace*{2.0em}\{\{"id": 2, "instruction": "...", "milestone": "..."\}\}\par
\noindent \hspace*{1.0em}]\par
\noindent \}\}\par
\noindent ```\par
\smallskip
\noindent Note: i0 does not have a milestone. Only i\_k (k >= 1) has a milestone.\par
\smallskip
\noindent Here are some examples:\par
\smallskip
\noindent **Example 1:**\par
\noindent Original instruction: `Find a hairless dog breed with medium barking.`\par
\noindent I0: `Find me a hairless dog breed.`\par
\smallskip
\noindent Constraints:\par
\noindent ```json\par
\noindent \{\{\par
\noindent \hspace*{1.0em}"constraints": [\par
\noindent \hspace*{2.0em}\{\{"id": 1, "description": "Must be a dog breed"\}\},\par
\noindent \hspace*{2.0em}\{\{"id": 2, "description": "Must be hairless"\}\},\par
\noindent \hspace*{2.0em}\{\{"id": 3, "description": "Barking level must be medium"\}\}\par
\noindent \hspace*{1.0em}]\par
\noindent \}\}\par
\noindent ```\par
\smallskip
\noindent ```json\par
\noindent \{\{\par
\noindent \hspace*{1.0em}"instructions": [\par
\noindent \hspace*{2.0em}\{\{"id": 0, "instruction": "Find me a hairless dog breed."\}\},\par
\noindent \hspace*{2.0em}\{\{"id": 1, "instruction": "Now only show breeds with medium barking.",\par
\noindent \hspace*{2.5em}"milestone": "The search context for hairless dog breeds is established."\}\}\par
\noindent \hspace*{1.0em}]\par
\noindent \}\}\par
\noindent ```\par
\smallskip
\noindent **Example 2:**\par
\noindent Original instruction: `I am a DC Comics fan. Could you find three comic books from different series written by the same author as Injustice: Gods Among Us Year One that feature Batman (excluding Injustice itself)? Please provide the links to the homepage of Injustice and these comic books on DC comic's official website.`\par
\noindent I0: `On DC.com, find Injustice: Gods Among Us Year One and other comic books by the same author.`\par
\smallskip
\noindent Constraints:\par
\noindent ```json\par
\noindent \{\{\par
\noindent \hspace*{1.0em}"constraints": [\par
\noindent \hspace*{2.0em}\{\{"id": 1, "description": "Comics must be found on DC.com"\}\},\par
\noindent \hspace*{2.0em}\{\{"id": 2, "description": "The comics must be written by the same author as Injustice: Gods Among Us Year One"\}\},\par
\noindent \hspace*{2.0em}\{\{"id": 3, "description": "Each of the comics kept after Year One must feature Batman"\}\},\par
\noindent \hspace*{2.0em}\{\{"id": 4, "description": "The final answer must identify exactly three comic books besides Year One"\}\},\par
\noindent \hspace*{2.0em}\{\{"id": 5, "description": "The three comic books must be from different series"\}\},\par
\noindent \hspace*{2.0em}\{\{"id": 6, "description": "Injustice: Gods Among Us Year One itself must be excluded from the three comic books"\}\},\par
\noindent \hspace*{2.0em}\{\{"id": 7, "description": "The final answer must include a DC.com page link for Injustice: Gods Among Us Year One"\}\},\par
\noindent \hspace*{2.0em}\{\{"id": 8, "description": "The final answer must include a DC.com page link for each of the three comic books"\}\}\par
\noindent \hspace*{1.0em}]\par
\noindent \}\}\par
\noindent ```\par
\smallskip
\noindent ```json\par
\noindent \{\{\par
\noindent \hspace*{1.0em}"instructions": [\par
\noindent \hspace*{2.0em}\{\{"id": 0, "instruction": "On DC.com, find Injustice: Gods Among Us Year One and other comic books by the same author."\}\},\par
\noindent \hspace*{2.0em}\{\{"id": 1, "instruction": "Only keep ones that feature Batman.",\par
\noindent \hspace*{2.5em}"milestone": "The DC.com search for Injustice: Gods Among Us Year One and other comic books by the same author is established."\}\},\par
\noindent \hspace*{2.0em}\{\{"id": 2, "instruction": "I want three of them, each from a different series, and not Injustice: Gods Among Us Year One itself.",\par
\noindent \hspace*{2.5em}"milestone": "The search for comics by the Year One author is being carried out under the Batman-feature criterion."\}\},\par
\noindent \hspace*{2.0em}\{\{"id": 3, "instruction": "Give the DC.com page link for Year One and for each of those three comics.",\par
\noindent \hspace*{2.5em}"milestone": "Three comics by that author, from different series and excluding Injustice itself, are being identified under the Batman criterion."\}\}\par
\noindent \hspace*{1.0em}]\par
\noindent \}\}\par
\noindent ```\par
\smallskip
\noindent Before you answer, self-check:\par
\noindent - Ensure each i\_k (k >= 1) milestone is a branch-robust semantic state derived only from prior instructions.\par
\noindent - Ensure JSON output follows the schema.\par
\smallskip
\noindent \{previous\_step4\_context\}\par
\smallskip
\noindent \{retry\_reflection\_context\}\par
\smallskip
\noindent Now design the multi-turn instruction sequence with milestones for the following:\par
\smallskip
\noindent Original instruction: \{original\_instruction\}\par
\smallskip
\noindent Constraints:\par
\noindent \{constraints\}\par
\smallskip
\noindent I0: \{i0\}\par
}\medskip

\end{prompt}

\subsubsection{Override Revision and Milestone Construction}
\begin{prompt}
\raggedright
\sloppy
\setlength{\emergencystretch}{3em}


\textbf{System}\par\smallskip
{\ttfamily
\noindent You are an expert at designing natural multi-turn user interactions for GUI tasks. Arrange extracted constraints into a realistic timeline, with a milestone for each turn after I0. Prefer a row from the candidate list as a reference. If that list is empty, propose a replacement for one existing constraint. You decide which preference to keep and what it changes to. Follow the override\_mode section exactly.\par
}\medskip


\textbf{User}\par\smallskip
{\ttfamily
\noindent Given the original task, constraints, I0, website, and override candidates, write a multi-turn instruction sequence with milestones.\par
\smallskip
\noindent **I0:** already provided \textemdash{} do not modify it.\par
\smallskip
\noindent **Later turns:**\par
\noindent - Apart from the selected override and its switching condition, use only the provided constraints. Additive turns may introduce remaining constraints, but must not add new requirements. Do not invent extra turns when no constraints remain.\par
\noindent - Sound like a user refining the request after seeing progress.\par
\noindent - Group information naturally; do not split one idea across turns.\par
\noindent - Keep composite AND/OR constraints intact.\par
\noindent - If `override\_candidates` is non-empty: pick exactly one row and copy `target\_constraint\_id` / `replacement\_pair\_id` into `selected`. Do not invent a pair. Write `keep` and `change\_to` yourself. `pair\_values` are a reference only; do not copy them unless that wording already fits the task.\par
\noindent - If `override\_candidates` is empty: pick one constraint as the preference the user could change (not the task subject). Write `keep` and `change\_to` on the same dimension. `change\_to` must be a different concrete value this website can actually show. Set `replacement\_pair\_id` to null. Do not change the subject (movies stay movies). Do not invent a price on a page with no prices.\par
\noindent - `"enable\_override": false` if no such preference exists, or if keep and change\_to would be the same preference.\par
\noindent - Do not rewrite a replacement into "any is fine" / "remove the filter" / "don't care".\par
\smallskip
\noindent \{mode\_block\}\par
\smallskip
\noindent **Style:** natural user voice ("Show me\ldots{}", "Only show\ldots{}", "Now set\ldots{}").\par
\smallskip
\noindent \#\# Milestones\par
\smallskip
\noindent Each i\_k (k >= 1) needs a milestone: a branch-robust semantic state after correctly completing i0..i\_\{\{k-1\}\}. User-visible progress, not UI clicks. Derive it only from prior turns; never leak i\_k. It must remain valid on empty-result branches. Prefer an earlier reachable state over restating i0's final goal.\par
\smallskip
\noindent Override insertion timing:\par
\noindent \{override\_timing\_guidance\}\par
\smallskip
\noindent \{previous\_step4\_context\}\par
\smallskip
\noindent Website: \{website\}\par
\smallskip
\noindent Original instruction: \{original\_instruction\}\par
\smallskip
\noindent Constraints:\par
\noindent \{constraints\}\par
\smallskip
\noindent I0: \{i0\}\par
\smallskip
\noindent Override candidates (each row is one seed constraint plus a reference pair; `pair\_values` are unlabeled; `pair\_dimension` is the pair's axis, not a switching quantity):\par
\noindent \{override\_candidates\}\par
\smallskip
\noindent Notes:\par
\noindent - i0 has no milestone.\par
\noindent - Turns after i0 need `type`.\par
\smallskip
\noindent Output JSON only.\par
\smallskip
}\medskip


\textbf{User: \{mode\_block\}, unconditional}\par\smallskip
{\ttfamily
\noindent \#\# Override mode: unconditional\par
\smallskip
\noindent Use a candidate row if any exist; otherwise use the empty-candidate rules. Write one `type="override"` turn that switches the current preference (`keep`) to `change\_to`. No if-clause. Other turns are `type="additive"`.\par
\smallskip
\noindent Candidate `pair\_values` are a reference on one dimension, not gold and not a direction label. Write `keep` and `change\_to` in the task's own wording. They may differ from the pair texts and from the anchor constraint.\par
\smallskip
\noindent When using a candidate, `selected` has:\par
\noindent - `target\_constraint\_id` / `replacement\_pair\_id` copied from that row\par
\noindent - `keep`: the preference that stays in force until the override\par
\noindent - `change\_to`: the preference after the override (a different concrete value)\par
\smallskip
\noindent Good: "Actually, change the color to blue instead. Keep the other criteria the same."\par
\noindent Bad: "Color doesn't matter anymore." / "If it's expensive, blue is fine."\par
\noindent Bad: keep and change\_to that are the same preference.\par
\smallskip
\noindent Example with candidates (duration 6\textendash{}12 min \ensuremath{\rightarrow} 12\textendash{}18 min). Copy real candidate ids; do not reuse the example ids.\par
\smallskip
\noindent ```json\par
\noindent \{\par
\noindent \hspace*{1.0em}"enable\_override": true,\par
\noindent \hspace*{1.0em}"selected": \{\par
\noindent \hspace*{2.0em}"target\_constraint\_id": 3,\par
\noindent \hspace*{2.0em}"replacement\_pair\_id": "from\_a\_candidate\_row",\par
\noindent \hspace*{2.0em}"keep": "The talk must last 6 to 12 minutes.",\par
\noindent \hspace*{2.0em}"change\_to": "The talk must last 12 to 18 minutes."\par
\noindent \hspace*{1.0em}\},\par
\noindent \hspace*{1.0em}"instructions": [\par
\noindent \hspace*{2.0em}\{"id": 0, "instruction": "Find me a TED talk on the topic of robots that lasts 6 to 12 minutes."\},\par
\noindent \hspace*{2.0em}\{"id": 1, "instruction": "Actually, change the duration to 12 to 18 minutes instead. Keep the topic the same, and pick the most viewed one.",\par
\noindent \hspace*{2.5em}"milestone": "The search context for TED talks on the topic of robots with a duration preference is established.",\par
\noindent \hspace*{2.5em}"type": "override"\}\par
\noindent \hspace*{1.0em}]\par
\noindent \}\par
\noindent ```\par
\smallskip
\noindent Empty-candidate example (IMDb).\par
\noindent Good: "Actually, use 2023 instead of 2024. Keep the matching China/US theatrical dates."\par
\noindent Bad: "Find TV shows instead." / "Any year is fine."\par
\smallskip
\noindent ```json\par
\noindent \{\par
\noindent \hspace*{1.0em}"enable\_override": true,\par
\noindent \hspace*{1.0em}"selected": \{\par
\noindent \hspace*{2.0em}"target\_constraint\_id": 2,\par
\noindent \hspace*{2.0em}"replacement\_pair\_id": null,\par
\noindent \hspace*{2.0em}"keep": "The movies must have premiered in 2024.",\par
\noindent \hspace*{2.0em}"change\_to": "The movies must have premiered in 2023.",\par
\noindent \hspace*{2.0em}"pair\_dimension": "year"\par
\noindent \hspace*{1.0em}\},\par
\noindent \hspace*{1.0em}"instructions": [\par
\noindent \hspace*{2.0em}\{"id": 0, "instruction": "On IMDb, search for movies premiered in 2024 that have matching theatrical release dates in mainland China and the United States."\},\par
\noindent \hspace*{2.0em}\{"id": 1, "instruction": "Actually, use 2023 instead of 2024. Keep the matching China/US theatrical dates.",\par
\noindent \hspace*{2.5em}"milestone": "The IMDb search for 2024 movies with matching China/US theatrical dates is established.",\par
\noindent \hspace*{2.5em}"type": "override"\}\par
\noindent \hspace*{1.0em}]\par
\noindent \}\par
\noindent ```\par
\smallskip
}\medskip


\textbf{User: \{mode\_block\}, conditional}\par\smallskip
{\ttfamily
\noindent \#\# Override mode: conditional\par
\smallskip
\noindent Use a candidate row if any exist; otherwise use the empty-candidate rules. The default stays `keep`. `change\_to` is accepted only when a page-observable condition holds.\par
\smallskip
\noindent A conditional override has two axes:\par
\smallskip
\noindent - Replacement dimension \textemdash{} the preference that changes between `keep` and `change\_to` (what the user is willing to revise: color, ticket type, distance, a rating bar, how many items, and so on).\par
\noindent - Condition dimension (`quantity`) \textemdash{} what is read from the page to decide whether the alternative is allowed. Choose it on its own. It may be the same as the replacement dimension, or different from it.\par
\smallskip
\noindent Condition dimension is one of:\par
\noindent - price  \textemdash{} a listed price or a price gap (include a currency)\par
\noindent - rating \textemdash{} stars, rank, or a similar score\par
\noindent - time   \textemdash{} duration, clock time, or recency on the page\par
\noindent - count  \textemdash{} how many results match a preference on the page\par
\smallskip
\noindent Choose a quantity this website can actually show. Write the condition only (not the replacement pair) in this form:\par
\smallskip
\noindent \hspace*{1.0em}(QUANTITY) ( > or < ) (NUMBER)\par
\smallskip
\noindent Those two axes combine in two natural ways:\par
\smallskip
\noindent 1. Advantage / cost \textemdash{} The condition compares the default outcome and the alternative outcome on price, rating, or time (for example, how much more the preferred option costs than the fallback). The replacement dimension is usually a categorical preference; the condition dimension is the tradeoff being weighed.\par
\smallskip
\noindent 2. Existence \textemdash{} The condition is that the results required under the default preference are not there: one matching result, or the K results the task asked for. The user then relaxes the preference, or relaxes how many are required. `count` is the usual quantity here.\par
\smallskip
\noindent Prefer a condition a real user would say after seeing results: a tradeoff they would weigh, or a shortfall they would notice. The numbers should come from the task and the page, so the reason for changing one's mind is easy to follow. Conditions that almost never fire, almost always fire, or only restate that nothing matched, rarely sound like real speech.\par
\smallskip
\noindent Do not restate the replacement pair as the condition itself.\par
\smallskip
\noindent Fill NUMBER from the task and the page.\par
\smallskip
\noindent The revision must keep `keep` as the default, state the condition with the concrete number, and accept `change\_to` only if the condition holds. `type="conditional\_override"`. Also set `condition\_text` on that turn: one predicate sentence using the same number and comparison.\par
\smallskip
\noindent Use one task-level condition to choose between `keep` and `change\_to` for the whole task when multiple entities are involved, explicitly state whether all or any must meet the condition, and apply the chosen preference uniformly rather than switching separately for each entity.\par
\smallskip
\noindent Candidate `pair\_values` are a reference, not gold and not a direction label. Write `keep` and `change\_to` in the task's own wording.\par
\smallskip
\noindent Milestone: the default preference has been applied, including empty or short lists.\par
\smallskip
\noindent If you cannot form a real preference change (from a candidate or a proposal), or no QUANTITY is observable on this page, return `"enable\_override": false`. Do not write a flat override.\par
\smallskip
\noindent `selected` fields:\par
\noindent - `target\_constraint\_id`\par
\noindent - `replacement\_pair\_id` copied from a candidate row, or null if proposing\par
\noindent - `keep` / `change\_to` (required; different preferences)\par
\noindent - if proposing: `pair\_dimension`\par
\noindent - `quantity`: price | rating | time | count\par
\noindent - `operator`: > or <\par
\noindent - `threshold`: the NUMBER\par
\noindent - `unit`: e.g. USD, or ""\par
\smallskip
\noindent Copy real candidate ids in your output; do not reuse the example ids.\par
\smallskip
\noindent Examples (keep \ensuremath{\rightarrow} change\_to, condition):\par
\smallskip
\noindent Advantage / cost:\par
\noindent - Replacement: non-red-eye \ensuremath{\leftrightarrow} red-eye. Condition: price gap.\par
\noindent \hspace*{1.0em}keep="Selected flight is non-red-eye", change\_to="Selected flight is red-eye",\par
\noindent \hspace*{1.0em}quantity=price, operator=">", threshold=500, unit="RMB"\par
\noindent \hspace*{1.0em}Good: "I prefer no red-eye flights. But if the cheapest non-red-eye flight is more than 500 RMB more expensive than the cheapest red-eye flight, red-eye is acceptable."\par
\noindent - Replacement: blue \ensuremath{\leftrightarrow} Space Gray. Condition: price gap.\par
\noindent \hspace*{1.0em}keep="Color must be blue", change\_to="Color must be Space Gray",\par
\noindent \hspace*{1.0em}quantity=price, operator=">", threshold=50, unit="USD"\par
\noindent \hspace*{1.0em}Good: "I still want blue. But if the cheapest blue is more than \$50 above the cheapest Space Gray, Space Gray is fine."\par
\smallskip
\noindent Existence:\par
\noindent - Replacement: must feature Batman \ensuremath{\leftrightarrow} must feature Superman. Condition: count.\par
\noindent \hspace*{1.0em}keep="Comics must feature Batman", change\_to="Comics must feature Superman",\par
\noindent \hspace*{1.0em}quantity=count, operator="<", threshold=2\par
\noindent \hspace*{1.0em}Good: "I want two comics featuring Batman. If fewer than two are listed, look for two comics featuring Superman instead."\par
\noindent - Replacement: need five \ensuremath{\rightarrow} accept three. Condition: count.\par
\noindent \hspace*{1.0em}keep="Find five matching items", change\_to="Three matching items are enough",\par
\noindent \hspace*{1.0em}quantity=count, operator="<", threshold=5\par
\noindent \hspace*{1.0em}Good: "I want five. If fewer than five are listed under that filter, three is fine."\par
\smallskip
\noindent Unnatural (users do not talk this way):\par
\noindent - "If there are fewer than 3 blue options, black is fine" when they only need one item. The task never asked for three; the 3 is invented, so the cutoff does not explain a real shortfall the user cares about.\par
\noindent - "If the cheapest blue is above \$500, accept Space Gray." Crossing an absolute dollar amount does not say why gray becomes acceptable \textemdash{} blue may still be the better deal than gray, and a real user would compare the two colors (or notice blue is missing), not switch color at a round price tag.\par
\smallskip
\noindent Full timeline:\par
\smallskip
\noindent ```json\par
\noindent \{\par
\noindent \hspace*{1.0em}"enable\_override": true,\par
\noindent \hspace*{1.0em}"selected": \{\par
\noindent \hspace*{2.0em}"target\_constraint\_id": 3,\par
\noindent \hspace*{2.0em}"replacement\_pair\_id": "from\_a\_candidate\_row",\par
\noindent \hspace*{2.0em}"keep": "Selected flight is non-red-eye.",\par
\noindent \hspace*{2.0em}"change\_to": "Selected flight is red-eye.",\par
\noindent \hspace*{2.0em}"quantity": "price",\par
\noindent \hspace*{2.0em}"operator": ">",\par
\noindent \hspace*{2.0em}"threshold": 500,\par
\noindent \hspace*{2.0em}"unit": "RMB"\par
\noindent \hspace*{1.0em}\},\par
\noindent \hspace*{1.0em}"instructions": [\par
\noindent \hspace*{2.0em}\{"id": 0, "instruction": "Search Ctrip for flights from Beijing to Shanghai departing the day after tomorrow."\},\par
\noindent \hspace*{2.0em}\{"id": 1, "instruction": "Only keep non-red-eye options.",\par
\noindent \hspace*{2.5em}"milestone": "Search results for Beijing\textendash{}Shanghai flights departing the day after tomorrow are displayed, including a no-results message if none are available.",\par
\noindent \hspace*{2.5em}"type": "additive"\},\par
\noindent \hspace*{2.0em}\{"id": 2, "instruction": "I prefer no red-eye flights. But if the cheapest non-red-eye flight is more than 500 RMB more expensive than the cheapest red-eye flight, red-eye is acceptable.",\par
\noindent \hspace*{2.5em}"milestone": "The page shows the results under the non-red-eye preference, including a no-results message if none are available.",\par
\noindent \hspace*{2.5em}"type": "conditional\_override",\par
\noindent \hspace*{2.5em}"condition\_text": "The cheapest non-red-eye Beijing\textendash{}Shanghai flight for the day after tomorrow is more than 500 RMB above the cheapest red-eye on Ctrip."\},\par
\noindent \hspace*{2.0em}\{"id": 3, "instruction": "Pick one matching flight and tell me the flight number and price.",\par
\noindent \hspace*{2.5em}"milestone": "The page shows the cheapest available price for each flight type, or indicates that one or both flight types are unavailable.",\par
\noindent \hspace*{2.5em}"type": "additive"\}\par
\noindent \hspace*{1.0em}]\par
\noindent \}\par
\noindent ```\par
\smallskip
}\medskip


\textbf{User: \{override\_timing\_guidance\}, early}\par\smallskip
{\ttfamily
\noindent Use EARLY mode: place the override revision immediately after the related preference is first introduced.\par
}\medskip


\textbf{User: \{override\_timing\_guidance\}, late}\par\smallskip
{\ttfamily
\noindent Use LATE mode: you may place the override revision at any natural later point in the timeline. Choose the timing freely to keep the interaction natural.\par
}\medskip

\end{prompt}

\subsubsection{Milestone Review}
\begin{prompt}
\raggedright
\sloppy
\setlength{\emergencystretch}{3em}


\textbf{System}\par\smallskip
{\ttfamily
\noindent You are a quality reviewer for milestones in dynamically revised Web-agent tasks. Review every milestone in the provided task and determine whether it is a valid trigger for delivering the corresponding user instruction.\par
}\medskip


\textbf{User}\par\smallskip
{\ttfamily
\noindent \#\# Task setup\par
\smallskip
\noindent This is a Web-agent task in which the user's goal can be revised during execution. The agent first receives an initial instruction i0. Each later instruction i\_k is paired with a milestone M\_k that determines when i\_k should be sent.\par
\smallskip
\noindent During execution, there is a trigger that checks whether the current browser page satisfies the next milestone. When M\_k is reached, i\_k is delivered to the agent before its next interaction step.\par
\smallskip
\noindent Therefore, when M\_k is checked, the agent has received only i0 through i\_\{\{k-1\}\}. Together, these earlier instructions define the cumulative goal currently given to the agent. Instruction i\_k and all later instructions are still unknown to the agent.\par
\smallskip
\noindent When an earlier instruction replaces a requirement, use the replacement when determining the current goal.\par
\smallskip
\noindent The later instructions are included in this review only so that you can detect whether a milestone improperly depends on requirements that have not yet been sent. Do not use them to justify a milestone.\par
\smallskip
\smallskip
\noindent \#\# What is a valid milestone?\par
\smallskip
\noindent A milestone is a semantic stage of the task that determines when the next user instruction is delivered.\par
\smallskip
\noindent For M\_k to be valid, successful execution of the cumulative goal defined by i0 through i\_\{\{k-1\}\} should naturally have to reach the stage described by M\_k. The milestone should follow from the goal already given to the agent, rather than from the upcoming instruction i\_k.\par
\smallskip
\noindent A milestone should describe task-semantic progress rather than one particular way of interacting with the website. Different successful agents may take different UI paths, visit different intermediate pages, or use different numbers of steps while still reaching the same semantic stage.\par
\smallskip
\noindent The milestone should describe that a semantic stage has been reached, rather than require that the stage produced a particular result. Reaching the relevant search stage must count even when no matching results are found. Do not require a successful match or a minimum result count before delivering the next instruction.\par
\smallskip
\noindent The milestone must also be recognizable by the trigger. The trigger determines whether M\_k has been reached using only the current browser page; it does not see the conversation, the agent's reasoning, action history, or final answer.\par
\smallskip
\smallskip
\noindent \#\# Review criteria\par
\smallskip
\noindent For each instruction after i0, summarize what the cumulative goal before i\_k requires in prior\_requirements and what must be true for M\_k to trigger in milestone\_requirements. Then check the following:\par
\smallskip
\noindent \#\#\# 1. Prior-goal grounding\par
\smallskip
\noindent Does every substantive requirement in M\_k follow from the cumulative goal already specified by i0 through i\_\{\{k-1\}\}?\par
\smallskip
\noindent Reject M\_k if it requires a search, comparison, selection, filtering operation, fallback, or other work that the cumulative goal does not ask for or require to complete. List each such requirement in unsupported\_requirements, whether it first appears in i\_k, appears in a later instruction, or never appears in the instructions.\par
\smallskip
\noindent Also reject a milestone that adds, narrows, strengthens, or changes an earlier requirement, such as the company, location, rating, quantity, duration, or comparison range. For example, "find a hotel" does not support a milestone requiring a five-star hotel, and "the cheapest Carnival cruise" does not mean the cheapest cruise across all companies.\par
\smallskip
\noindent Judge semantic support rather than exact wording. A milestone may describe an intermediate search, comparison, or selection stage without those exact words appearing in the instructions, as long as reaching that stage is required to accomplish the cumulative goal.\par
\smallskip
\smallskip
\noindent \#\#\# 2. Inevitable semantic stage\par
\smallskip
\noindent Does successful execution of the cumulative goal require reaching the semantic stage described by M\_k?\par
\smallskip
\noindent Reject a milestone that describes only one possible way of completing the task. For example, if the task can be completed from the results page, a milestone should not require opening an individual result, because successful execution does not necessarily require that step.\par
\smallskip
\noindent The milestone should capture that the relevant stage of the task has been reached, regardless of the particular outcome produced at that stage. For example, reaching the relevant search-results stage may be valid even if the search returns no matching candidates.\par
\smallskip
\smallskip
\noindent \#\#\# 3. Path independence\par
\smallskip
\noindent Does M\_k describe semantic task progress rather than a particular UI path?\par
\smallskip
\noindent Reject a milestone that depends on a specific click, page transition, opened result, or sequence of UI operations when another successful execution of the same cumulative goal could avoid that state.\par
\smallskip
\smallskip
\noindent \#\#\# 4. Page observability\par
\smallskip
\noindent Can the trigger determine from the current browser page alone whether M\_k has been reached?\par
\smallskip
\noindent A page displaying the relevant search results or task information can provide evidence. Statements such as "the user has asked for alternatives" or "the agent is preparing an answer" cannot be verified from the current page alone.\par
\smallskip
\noindent Set observable to false when an essential part of the milestone requires evidence that is unavailable from the current page.\par
\smallskip
\smallskip
\noindent \#\#\# 5. Conditional revisions\par
\smallskip
\noindent If i\_k is a conditional revision, M\_k must not presuppose either outcome of the condition. The same milestone should be able to trigger regardless of which branch of the condition applies.\par
\smallskip
\noindent For example, if i\_k says to change the search when fewer than two results are found, M\_k must not require at least two results. Otherwise the revision would be blocked precisely when fewer than two results are found.\par
\smallskip
\noindent Likewise, M\_k must not require work introduced only by the conditional revision itself. Such work should already be rejected by the prior-goal grounding check.\par
\smallskip
\smallskip
\noindent \#\# Examples\par
\smallskip
\noindent \#\#\# Future-dependency failure\par
\smallskip
\noindent i0: Find the cheapest cruise departing from Los Angeles that lasts at least 8 days.\par
\smallskip
\noindent i1 [conditional\_override]: I still want to start with a Los Angeles cruise of at least 8 days. But if the cheapest 8-day Los Angeles sailing is more than \$500 above the cheapest Los Angeles sailing of 5 days or less, then a shorter cruise is fine. Compare cheapest against cheapest across all the Los Angeles sailings under each duration \textemdash{} same rule for the whole search, not ship by ship.\par
\smallskip
\noindent Milestone checked before sending i1:\par
\noindent Search results for Los Angeles cruises lasting at least 8 days are available, and a shorter-sailing fallback is available if the price gap is large enough.\par
\smallskip
\noindent Judgment: Fail. i0 requires finding the cheapest Los Angeles cruise lasting at least 8 days. It does not require a shorter-sailing fallback or a price-gap comparison. Those requirements are introduced only by i1, so the milestone depends on work that has not yet been requested.\par
\smallskip
\smallskip
\noindent \#\#\# Observability failure\par
\smallskip
\noindent i0: Search the Cleveland Museum of Art's collection for Chinese objects dated 2000 BCE or earlier.\par
\smallskip
\noindent i1 [additive]: From the museum's official website, identify exactly five of those objects.\par
\smallskip
\noindent i2 [additive]: For each of the five, give me its name, period, and estimated date range.\par
\smallskip
\noindent Milestone checked before sending i2:\par
\noindent The search context for Chinese objects from the Cleveland Museum of Art dated 2000 BCE or earlier is established, and the user has asked for exactly five objects from the museum's official website.\par
\smallskip
\noindent Judgment: Fail; observable is false. The requirement to identify exactly five objects is supported by i1, so it is part of the cumulative goal before i2. However, the current browser page cannot establish that the user made this request. This is an observability problem, not a future-dependency problem.\par
\smallskip
\noindent \#\# Pass and retry rules\par
\smallskip
\noindent A round passes only if all five checks above are satisfied. The overall review passes only if every round passes.\par
\smallskip
\noindent If the overall review fails, set worth\_retry to true if rewriting the instructions or milestones can resolve the problem while retaining the task. Set retry\_from\_step to "step2" if i0 must change; otherwise use "step3" to rewrite the later instructions or milestones.\par
\smallskip
\noindent If no such rewrite can resolve the problem, set worth\_retry to false and retry\_from\_step to null, and explain why. Leave both retry fields null when the overall review passes.\par
\smallskip
\noindent \#\# Output format\par
\smallskip
\noindent Return one check for every instruction after i0. The observable and passed fields must be JSON booleans (true or false). Use this JSON structure:\par
\noindent \{\{\par
\noindent \hspace*{1.0em}"milestone\_checks": [\par
\noindent \hspace*{2.0em}\{\{\par
\noindent \hspace*{3.0em}"instruction\_id": "i1",\par
\noindent \hspace*{3.0em}"prior\_requirements": "What the cumulative goal requires before this instruction",\par
\noindent \hspace*{3.0em}"milestone\_requirements": "What must be true for this milestone to trigger",\par
\noindent \hspace*{3.0em}"unsupported\_requirements": [],\par
\noindent \hspace*{3.0em}"observable": true,\par
\noindent \hspace*{3.0em}"passed": true,\par
\noindent \hspace*{3.0em}"reason": "A brief explanation of the judgment"\par
\noindent \hspace*{2.0em}\}\}\par
\noindent \hspace*{1.0em}],\par
\noindent \hspace*{1.0em}"passed": true,\par
\noindent \hspace*{1.0em}"reason": "A brief summary of the review",\par
\noindent \hspace*{1.0em}"worth\_retry": null,\par
\noindent \hspace*{1.0em}"retry\_from\_step": null\par
\noindent \}\}\par
\smallskip
\noindent In each reason, identify the milestone requirement and the earlier goal that supports it, or explain what prior support or page evidence is missing. If the milestone fails because it is not an inevitable semantic stage, depends on a particular execution path, or presupposes an outcome of a conditional revision, state that explicitly.\par
\smallskip
\noindent Instructions and milestones:\par
\noindent \{redesign\}\par
\smallskip
}\medskip

\end{prompt}

\subsubsection{Instruction Review}
\begin{prompt}
\raggedright
\sloppy
\setlength{\emergencystretch}{3em}


\textbf{System}\par\smallskip
{\ttfamily
\noindent You are a quality reviewer for user instruction sequences in dynamically revised Web-agent tasks. Review whether the rewritten instructions preserve the intended task while forming a clear, coherent, and natural user interaction.\par
}\medskip


\textbf{User}\par\smallskip
{\ttfamily
\noindent \#\# Task setup\par
\smallskip
\noindent This is a Web-agent task in which the user's goal can be revised during execution. The agent first receives an initial instruction i0, followed by later instructions i1, i2, and so on.\par
\smallskip
\noindent The agent receives these instructions in order. Requirements may be introduced gradually, but the resulting sequence should still resemble a natural user interaction rather than an arbitrary decomposition of the original task.\par
\smallskip
\noindent The original task defines the requirements that should ultimately be represented by the instruction sequence. A declared override may intentionally replace one of those requirements with a different requirement.\par
\smallskip
\noindent Review only the user instructions and their evolution. Milestone timing and milestone validity are checked separately.\par
\smallskip
\smallskip
\noindent \#\# What is a valid instruction sequence?\par
\smallskip
\noindent A valid sequence should preserve the intended task while expressing it as a plausible interaction between a user and an agent.\par
\smallskip
\noindent The initial instruction should give the agent a coherent and executable starting goal. Later instructions may add requirements, change preferences, relax earlier requirements, or introduce conditional changes.\par
\smallskip
\noindent A later requirement does not need to have been implied by the earlier instructions. Users are allowed to add information or change their mind during execution. However, the sequence should not look as though a single natural request was mechanically split into multiple turns only to create more revisions.\par
\smallskip
\noindent Each instruction must also be understandable in the context of the instructions already received. Later instructions may refer back to earlier entities or requirements, but those references must have a clear meaning.\par
\smallskip
\smallskip
\noindent \#\# Review criteria\par
\smallskip
\noindent \#\#\# 1. Clarity and executability\par
\smallskip
\noindent Can the agent tell what to do after each instruction using the requests received so far?\par
\smallskip
\noindent The initial instruction should specify a coherent task that the agent can meaningfully begin executing. Later instructions should be understandable in the context established by the earlier ones.\par
\smallskip
\noindent References such as "those options", "that hotel", or "the cheaper one" must have an identifiable meaning. For comparisons such as "cheaper", "nearer", or "higher-rated", the instructions must make clear what is being compared and what reference price, location, rating, or other quantity is relevant.\par
\smallskip
\noindent Flag requirements that cannot be satisfied together unless a later instruction explicitly replaces the earlier one.\par
\smallskip
\smallskip
\noindent \#\#\# 2. Faithfulness to the original task\par
\smallskip
\noindent Compare the complete instruction sequence with the original task.\par
\smallskip
\noindent Every original requirement should be introduced somewhere in the sequence unless it is replaced by the declared override. Preserve important qualifiers, scopes, comparison rules, and boundary conditions rather than weakening them through paraphrasing.\par
\smallskip
\noindent Any new requirement absent from the original task must come from the declared override or be needed to express its condition. A requirement should not be added merely because it is plausible or useful.\par
\smallskip
\noindent For example, rewriting "find a hotel" as "find a five-star hotel" adds a rating requirement unless that change is explicitly declared.\par
\smallskip
\smallskip
\noindent \#\#\# 3. Natural interaction structure\par
\smallskip
\noindent Does the sequence resemble a plausible way for a user's goal to evolve during interaction?\par
\smallskip
\noindent The initial instruction should not feel artificially incomplete merely because requirements have been withheld for later rounds. At the same time, it does not need to include every eventual preference.\par
\smallskip
\noindent Each later instruction should introduce a meaningful addition, change, relaxation, or conditional preference. Reject a revision that is only a trivial fragment of something a user would naturally state together with the preceding instruction.\par
\smallskip
\noindent Closely related requirements that would normally be expressed in the same user turn should not be mechanically separated only to increase the number of revisions.\par
\smallskip
\noindent Do not reject a revision merely because the user could have stated it earlier. Users may naturally add requirements or reconsider preferences during execution. The issue is whether the resulting interaction itself feels artificial or incoherent.\par
\smallskip
\noindent Also consider the sequence as a whole. Later instructions should continue the same underlying task unless they explicitly revise part of it, and the ordering of revisions should make sense as a progression of user intent.\par
\smallskip
\smallskip
\noindent \#\#\# 4. Override validity\par
\smallskip
\noindent If there is an override, identify the earlier requirement it replaces and the replacement requirement. Restating the same requirement is not an override. Check that the requirement being replaced and its replacement match the declared override.\par
\smallskip
\noindent The replacement should revise a requirement that can meaningfully change while preserving the underlying task. It should not ambiguously redefine the task or leave it unclear which requirement is currently active.\par
\smallskip
\noindent For a conditional override, check that the default requirement appears before the conditional instruction, that the instruction clearly states when to retain it and when to use the replacement, and that the two branches agree with the declared override.\par
\smallskip
\noindent Identify the count, price, rating, time, distance, or other quantity being compared, its threshold, and whether the boundary is inclusive when that distinction matters. For example, "fewer than two" differs from "at most two".\par
\smallskip
\noindent The condition must be decidable from information the agent can obtain while carrying out the task. It must select one branch for the whole task rather than a different branch independently for each item.\par
\smallskip
\smallskip
\noindent \#\# Examples\par
\smallskip
\noindent \#\#\# Ambiguous replacement requirement\par
\smallskip
\noindent Original task:\par
\noindent Find the top-rated hotel in Manhattan, NY, suitable for 4 guests, and identify the fastest public transportation option from the hotel to LGA airport.\par
\smallskip
\noindent Instructions:\par
\noindent i0: Find a hotel in Manhattan, NY that is suitable for 4 guests.\par
\noindent i1 [additive]: Now narrow the results to the top-rated hotel among those options.\par
\noindent i2 [override]: Actually, change the hotel rating requirement from top-rated to 5-star reviews. Keep Manhattan and 4 guests.\par
\noindent i3 [additive]: For that hotel, find the fastest public transportation option from the hotel to LGA airport.\par
\smallskip
\noindent Judgment: Fail. In i2, "5-star reviews" does not specify whether the hotel must have an average review score of 5.0 or at least one five-star review. These select different hotels. A hotel's star classification is also different from its review score. Even when replacing the rating requirement is allowed, the replacement must specify which measure it uses.\par
\smallskip
\smallskip
\noindent \#\#\# Omitted original qualifier\par
\smallskip
\noindent Original task:\par
\noindent Identify five ancient Chinese objects from the Cleveland Museum of Art's collection that dated entirely to 2000 BCE or earlier, as listed on the museum's official website. For each object, provide its name, period, estimated date range, overall dimensions, and current location within the museum. If an object is not currently on display, please indicate that as well.\par
\smallskip
\noindent Instructions:\par
\noindent i0: Search the Cleveland Museum of Art's collection for Chinese objects dated 2000 BCE or earlier.\par
\noindent i1 [additive]: From the museum's official website, identify exactly five of those objects.\par
\noindent i2 [additive]: For each of the five, give me its name, period, and estimated date range.\par
\noindent i3 [additive]: Also include each object's overall dimensions and its current location within the museum.\par
\noindent i4 [additive]: If any object is not currently on display, please note that it is not on display.\par
\noindent i5 [override]: Actually, change the number to exactly three objects instead of five. Keep the official website, Chinese, 2000 BCE or earlier criteria and all the same details.\par
\smallskip
\noindent Judgment: Fail. The original task requires each object's entire date range to be 2000 BCE or earlier. The rewritten sequence says "dated 2000 BCE or earlier" and never states that the entire range must satisfy the cutoff. For an object dated 2100-1900 BCE, the range crosses that cutoff. The instructions must preserve the original requirement that the entire range qualify; changing the number of objects in i5 does not replace the date requirement.\par
\smallskip
\smallskip
\noindent \#\# Pass and retry rules\par
\smallskip
\noindent Set passed to true only if all applicable checks above pass.\par
\smallskip
\noindent If there is a problem, identify the relevant instruction and wording in reason, and explain whether the problem is ambiguity, contradiction, missing or added requirements, an invalid override, or an unnatural interaction structure.\par
\smallskip
\noindent If the review fails, set worth\_retry to true if rewriting the instructions can resolve the problem while retaining the task. Set retry\_from\_step to "step2" if i0 must change; otherwise use "step3" to rewrite the later instructions.\par
\smallskip
\noindent If no such rewrite can resolve the problem, set worth\_retry to false and retry\_from\_step to null, and explain why. Leave both retry fields null when the review passes.\par
\smallskip
\smallskip
\noindent \#\# Output format\par
\smallskip
\noindent Return your judgment using this JSON structure:\par
\noindent \{\{\par
\noindent \hspace*{1.0em}"passed": true,\par
\noindent \hspace*{1.0em}"reason": "A brief explanation, identifying the relevant instructions if there is a problem",\par
\noindent \hspace*{1.0em}"worth\_retry": null,\par
\noindent \hspace*{1.0em}"retry\_from\_step": null\par
\noindent \}\}\par
\smallskip
\noindent The passed field must be a JSON boolean (true or false).\par
\smallskip
\smallskip
\noindent Original task:\par
\noindent \{original\_instruction\}\par
\smallskip
\noindent Declared override (if any):\par
\noindent \{override\_context\}\par
\smallskip
\noindent Instructions:\par
\noindent \{instructions\}\par
\smallskip
}\medskip

\end{prompt}

\subsubsection{Constraint Synchronization}
\begin{prompt}
\raggedright
\sloppy
\setlength{\emergencystretch}{3em}


\textbf{User: unconditional}\par\smallskip
{\ttfamily
\noindent This is the whole task:\par
\smallskip
\noindent \{instructions\}\par
\smallskip
\noindent The user's requirements change during the task. A revision in a later turn may override an earlier requirement, or select according to a condition, and that decides some requirements of the final goal.\par
\smallskip
\noindent There are several constraints describing the requirements of the final goal. Each constraint is independent of the others, and together they describe the final goal.\par
\smallskip
\noindent \{constraints\}\par
\smallskip
\noindent Because the user's requirements changed, one constraint has already been updated correctly:\par
\noindent - id: \{target\_id\}\par
\noindent - before: \{keep\}\par
\noindent - now: \{now\}\par
\smallskip
\noindent Other constraints may still contain related wording that was not updated. Find those descriptions and change them so they agree with this updated constraint.\par
\smallskip
\noindent Return JSON:\par
\noindent \{\{"edits": [\{\{"id": <int>, "description": "<updated sentence>"\}\}]\}\}\par
\smallskip
\noindent Example:\par
\smallskip
\noindent This is the whole task:\par
\noindent i0: Find a certified refurbished iPad Air in blue with 256GB storage.\par
\noindent i1: Pick the cheapest one.\par
\noindent i2: Change the storage to 512GB. Keep the blue color and the cheapest pick.\par
\smallskip
\noindent Constraints:\par
\noindent 1. Color must be blue.\par
\noindent 2. Storage capacity must be 512GB.\par
\noindent 3. The selected product must be the cheapest among certified refurbished iPad Air models with 256GB storage in blue.\par
\smallskip
\noindent One constraint has already been updated correctly:\par
\noindent - id: 2\par
\noindent - before: Storage capacity must be 256GB.\par
\noindent - now: Storage capacity must be 512GB.\par
\smallskip
\noindent \{\{"edits": [\{\{"id": 3, "description": "The selected product must be the cheapest among certified refurbished iPad Air models with 512GB storage in blue."\}\}]\}\}\par
\smallskip
\noindent Now do this for the task above.\par
\smallskip
}\medskip


\textbf{User: conditional}\par\smallskip
{\ttfamily
\noindent This is the whole task:\par
\smallskip
\noindent \{instructions\}\par
\smallskip
\noindent The user's requirements change during the task. A revision in a later turn may override an earlier requirement, or select according to a condition, and that decides some requirements of the final goal.\par
\smallskip
\noindent There are several constraints describing the requirements of the final goal. Each constraint is independent of the others, and together they describe the final goal.\par
\smallskip
\noindent \{constraints\}\par
\smallskip
\noindent Because the user's requirements changed, one constraint is decided by a condition. It has two outcomes:\par
\noindent - id: \{target\_id\}\par
\noindent - default: \{keep\}\par
\noindent - alternative: \{change\_to\}\par
\noindent - condition: \{condition\}\par
\smallskip
\noindent Other constraints may still restate the default outcome only. For each of those, write two sentences: one for the default outcome, one for the alternative. Do not merge them into one sentence with if.\par
\smallskip
\noindent Return JSON:\par
\noindent \{\{"splits": [\{\{"id": <int>, "default": "<sentence>", "alternative": "<sentence>"\}\}]\}\}\par
\smallskip
\noindent Example:\par
\smallskip
\noindent This is the whole task:\par
\noindent i0: Find a certified refurbished iPad Air in blue with 256GB storage.\par
\noindent i1: I still prefer blue. But if fewer than 1 blue option is listed, Space Gray is acceptable.\par
\noindent i2: Pick the cheapest one under that color rule.\par
\smallskip
\noindent Constraints:\par
\noindent 1. Color must be blue.\par
\noindent 2. Storage capacity must be 256GB.\par
\noindent 3. The selected product must be the cheapest among certified refurbished iPad Air models with 256GB storage in blue.\par
\smallskip
\noindent One constraint is decided by a condition:\par
\noindent - id: 1\par
\noindent - default: Color must be blue.\par
\noindent - alternative: Color may be Space Gray.\par
\noindent - condition: fewer than 1 blue certified refurbished iPad Air with 256GB is listed.\par
\smallskip
\noindent \{\{"splits": [\{\{"id": 3, "default": "The selected product must be the cheapest among certified refurbished iPad Air models with 256GB storage in blue.", "alternative": "The selected product must be the cheapest among certified refurbished iPad Air models with 256GB storage in Space Gray."\}\}]\}\}\par
\smallskip
\noindent Now do this for the task above.\par
\smallskip
}\medskip

\end{prompt}

\subsubsection{Constraint Binding}
\begin{prompt}
\raggedright
\sloppy
\setlength{\emergencystretch}{3em}


\textbf{System}\par\smallskip
{\ttfamily
\noindent You are an expert at analyzing multi-turn task designs and mapping constraints to instructions. Your job is to determine which instruction introduces or covers each constraint, and to audit the coverage quality.\par
}\medskip


\textbf{User}\par\smallskip
{\ttfamily
\noindent You are given:\par
\noindent 1. A list of constraints extracted from a task\par
\noindent 2. The I0 (initial instruction)\par
\noindent 3. The full multi-turn instruction sequence (including I0)\par
\smallskip
\noindent Your job is to bind each constraint to the instruction that introduces or covers it, and audit the quality of constraint coverage.\par
\smallskip
\noindent \{override\_context\_block\}\par
\smallskip
\noindent **Rules:**\par
\noindent - Each constraint must be mapped to exactly one instruction (the one that most directly introduces or addresses it)\par
\noindent - Use instruction IDs: "i0" for the initial instruction, "i1", "i2", etc. for subsequent ones\par
\noindent - Rate your confidence for each mapping: "high", "medium", or "low"\par
\noindent - Provide a brief reason for each mapping\par
\smallskip
\noindent **Audit checks:**\par
\noindent - Flag any constraint not clearly covered by any instruction\par
\noindent - Flag any constraint that seems poorly matched to its assigned instruction\par
\noindent - Flag if the overall coverage is incomplete (constraints missing from the instruction sequence)\par
\noindent - Also check that every requirement in the instruction sequence is covered by a scoring constraint. Flag any extra requirement without coverage.\par
\noindent - If override context is provided, allow the target constraint to be satisfied via an explicit target->replacement override instruction.\par
\smallskip
\noindent Constraints:\par
\noindent \{constraints\}\par
\smallskip
\noindent I0: \{i0\}\par
\smallskip
\noindent Full instruction sequence:\par
\noindent \{instructions\}\par
\smallskip
\noindent Output format:\par
\noindent ```json\par
\noindent \{\{\par
\noindent \hspace*{1.0em}"mapping": [\par
\noindent \hspace*{2.0em}\{\{"constraint\_id": 1, "instruction\_id": "i0", "confidence": "high", "reason": "I0 explicitly mentions ..."\}\},\par
\noindent \hspace*{2.0em}\{\{"constraint\_id": 2, "instruction\_id": "i1", "confidence": "medium", "reason": "i1 addresses this through ..."\}\}\par
\noindent \hspace*{1.0em}],\par
\noindent \hspace*{1.0em}"issues": [],\par
\noindent \hspace*{1.0em}"passed": true,\par
\noindent \hspace*{1.0em}"retry\_from\_step": null\par
\noindent \}\}\par
\noindent ```\par
\smallskip
\noindent Set "passed" to false if there are serious coverage gaps (constraints with no matching instruction, or multiple low-confidence mappings). Minor issues (e.g., one medium-confidence mapping) should still pass.\par
\smallskip
\noindent **When "passed" is false, set "retry\_from\_step":**\par
\noindent - "step2": The issue originates from I0 design \textemdash{} I0 is too vague or too narrow, causing downstream instructions to miss constraints or overlap awkwardly. Redo from I0 design.\par
\noindent - "step3": The issue is in how constraints were arranged across instructions \textemdash{} poor distribution, missing coverage, or redundant overlap. I0 itself is fine; redo the multi-turn arrangement.\par
\smallskip
\noindent Now analyze the constraint binding.\par
}\medskip


\textbf{User: \{override\_context\_block\}, conditional}\par\smallskip
{\ttfamily
\noindent Override context:\par
\noindent \{ctx\_json\}\par
\smallskip
\noindent This is a conditional override (override\_mode=conditional). Constraints may carry branch=default or branch=alternative.\par
\noindent - Bind the target default constraint and every branch=alternative constraint to the conditional\_override revision.\par
\noindent - Bind other branch=default constraints to the instruction that introduces that requirement (e.g. cheapest stays on the turn that asks for cheapest).\par
}\medskip


\textbf{User: \{override\_context\_block\}, unconditional}\par\smallskip
{\ttfamily
\noindent Override context:\par
\noindent \{ctx\_json\}\par
\smallskip
\noindent Binding rule: for any override item above, it is valid to bind the target constraint to the override instruction that explicitly replaces target with replacement.\par
}\medskip

\end{prompt}

\subsubsection{Construction Retry and Reflection}
\begin{prompt}
\raggedright
\sloppy
\setlength{\emergencystretch}{3em}


\textbf{User: \{retry\_reflection\_context\}}\par\smallskip
{\ttfamily
\noindent \#\# Retry reflection (REQUIRED before producing new design)\par
\smallskip
\noindent You have attempted this task \{n\_attempts\} time(s) before. The full history of prior attempts and reviewer critiques is below. Before producing a new design, you MUST first reflect on the history and output a diagnosis.\par
\smallskip
\noindent \{history\_block\}\par
\smallskip
\noindent Reflection questions (answer each honestly in the `reflection` field of your JSON output):\par
\smallskip
\noindent 1. **Common pattern**: What is the common thread across the reviewer's critiques? Are they    pointing to the same i\_k, the same milestone type (necessity / branch-neutrality /    content-leakage / UI-assumption), or different failures each time?\par
\smallskip
\noindent 2. **Dimensions already tried**: Across prior attempts, what dimensions of the design have    actually changed (e.g., milestone wording only, number of turns, constraint grouping,    which turn a constraint lives in, I0 scope)? Be specific \textemdash{} look at the diffs, not    intentions.\par
\smallskip
\noindent 3. **Stuck-in-a-loop check**: If prior attempts varied only milestone wording but the    critique kept pointing at the same semantic issue, you are in a text-swap loop. Say so    explicitly if this is the case.\par
\smallskip
\noindent 4. **Next move**: Based on 1-3, what dimension should this attempt change, and why? This    is YOUR call \textemdash{} do not default to either "small edit" or "big restructure". Let the    pattern you identified drive the choice. It is equally valid to conclude "prior attempts    varied structure unnecessarily; the fix is precise milestone rewording" OR "prior    attempts shuffled words; the fix is a different constraint grouping".\par
\smallskip
\noindent Output this reflection as a structured field in your JSON (see schema below). Then produce the new design per the rules above, with the chosen dimension actually changed.\par
\smallskip
\noindent Output format for retry:\par
\noindent ```json\par
\noindent \{\{\par
\noindent \hspace*{1.0em}"reflection": \{\{\par
\noindent \hspace*{2.0em}"common\_pattern": "...",\par
\noindent \hspace*{2.0em}"dimensions\_already\_tried": ["..."],\par
\noindent \hspace*{2.0em}"in\_text\_swap\_loop": true or false,\par
\noindent \hspace*{2.0em}"next\_move": "...",\par
\noindent \hspace*{2.0em}"rationale": "..."\par
\noindent \hspace*{1.0em}\}\},\par
\noindent \hspace*{1.0em}"instructions": [\par
\noindent \hspace*{2.0em}\{\{"id": 0, "instruction": "..."\}\},\par
\noindent \hspace*{2.0em}\{\{"id": 1, "instruction": "...", "milestone": "..."\}\}\par
\noindent \hspace*{1.0em}]\par
\noindent \}\}\par
\noindent ```\par
\smallskip
}\medskip


\textbf{User: I0 review feedback}\par\smallskip
{\ttfamily
\noindent **Previous attempt feedback:** \{previous\_step4\_reason\}\par
\smallskip
\noindent Please address the issues identified above in your redesign of I0.\par
}\medskip


\textbf{User: JSON-format retry suffix}\par\smallskip
{\ttfamily
\smallskip
\smallskip
\noindent IMPORTANT: Return ONLY a valid JSON object matching the schema. Do not include explanations, headings, or markdown.\par
}\medskip


\textbf{User: Additive review feedback}\par\smallskip
{\ttfamily
\noindent **Previous attempt feedback:** \{previous\_step4\_reason\}\par
\smallskip
\noindent Please address the issues identified above in your redesign of the multi-turn sequence.\par
}\medskip


\textbf{User: I0-only retry suffix}\par\smallskip
{\ttfamily
\smallskip
\smallskip
\noindent Your previous output is invalid because it only contains i0. Regenerate and include at least i1 with a valid milestone. Return JSON only.\par
}\medskip


\textbf{User: Override review feedback}\par\smallskip
{\ttfamily
\noindent Previous review feedback:\par
\noindent \{previous\_step4\_reason\}\par
\noindent Regenerate while fixing the above issues.\par
}\medskip

\end{prompt}

\subsection{Agent Execution and Revision Delivery}

\subsubsection{Policy Agent}
\begin{prompt}
\raggedright
\sloppy
\setlength{\emergencystretch}{3em}


\textbf{System}\par\smallskip
{\ttfamily
\noindent You are an autonomous browser-control agent.\par
\smallskip
\noindent PAGE CONTENT shows the accessibility tree. Each interactive element is shown as:\par
\noindent \hspace*{1.0em}[bid] role 'text'\par
\noindent where bid is a numeric ID (e.g. 173), role is the element type, and 'text' is the visible label.\par
\smallskip
\noindent Output contract (strict):\par
\noindent - The only executable output is one markdown fence labeled action.\par
\noindent - The fence body must contain exactly one function call and nothing else.\par
\noindent - Text outside the fence is ignored. Never put an executable call outside the fence.\par
\noindent - Never use XML, DSML, JSON, unlabeled fences, or a bare last-line call.\par
\noindent - click, fill, and press require the NUMERIC bid from the [bid] tag, NOT the element text.\par
\noindent \hspace*{1.0em}Example: for [173] link 'Skip to main content', use click("173"), NOT click("Skip to main content").\par
\noindent - Never ask user for extra input.\par
\smallskip
\noindent Required layout (copy this shape exactly):\par
\smallskip
\noindent ```action\par
\noindent click("173")\par
\noindent ```\par
\smallskip
\noindent Valid actions and their signatures:\par
\noindent \hspace*{2.0em}click("bid")              \textendash{} click element by its numeric bid\par
\noindent \hspace*{2.0em}fill("bid", "value")      \textendash{} type text into an input\par
\noindent \hspace*{2.0em}press("bid", "key\_comb")  \textendash{} press key(s) on a focused element\par
\noindent \hspace*{2.0em}scroll(delta\_x, delta\_y)  \textendash{} scroll in pixels (positive = right/down)\par
\noindent \hspace*{2.0em}goto("url")               \textendash{} navigate to a URL\par
\noindent \hspace*{2.0em}send\_msg\_to\_user("text")  \textendash{} send message to the user\par
\noindent \hspace*{2.0em}report\_infeasible("reason")\par
\smallskip
\noindent Wrong (these are discarded): a bare `click("173")` on the last line; DSML/XML\par
\noindent tool markup; JSON; extra calls inside the fence; an unlabeled ``` fence.\par
\smallskip
\noindent Rules:\par
\noindent 1) Stay on the target website whenever possible.\par
\noindent 2) Follow ALL current instructions (initial + injected revisions).\par
\noindent 3) For conditional instructions, read page evidence before deciding.\par
\noindent 4) When you finish the task, put send\_msg\_to\_user("done: ...") inside the\par
\noindent \hspace*{1.5em}action fence. Clearly state in the message what you accomplished, with\par
\noindent \hspace*{1.5em}concrete evidence \textemdash{} product names, links, prices, or other specific facts\par
\noindent \hspace*{1.5em}that demonstrate completion.\par
\smallskip
\noindent \hspace*{1.5em}If a turn contained a conditional clause (e.g. "if X, then Y, otherwise\par
\noindent \hspace*{1.5em}keep Z") \textemdash{} typically marked with [CONDITIONAL] in the instruction list \textemdash{}\par
\noindent \hspace*{1.5em}include a [CONDITIONAL OBSERVATION] section in your done\_message that:\par
\noindent \hspace*{1.5em}- states what you observed for the condition on the page (with concrete\par
\noindent \hspace*{2.5em}values: prices, counts, ratings, distances, etc.),\par
\noindent \hspace*{1.5em}- reports whether the condition holds based on that observation,\par
\noindent \hspace*{1.5em}- identifies which branch (default or alternative) you took as a result.\par
\noindent 5) Explore thoroughly before giving up. If the current page does not show what you need, try navigating, searching, scrolling, or visiting related URLs. Exhaust all reasonable paths first.\par
\smallskip
\noindent Note: PAGE CONTENT shows only elements currently visible in the browser viewport (1280x720).\par
\noindent Content above or below the visible area is not shown. Use scroll(0, 300) to scroll down\par
\noindent or scroll(0, -300) to scroll up to reveal more content.\par
}\medskip

\end{prompt}

\subsubsection{Per-step Input Template}
\begin{prompt}
\raggedright
\sloppy
\setlength{\emergencystretch}{3em}


\textbf{User}\par\smallskip
{\ttfamily
\noindent CURRENT INSTRUCTIONS (ALL active):\par
\noindent \{snapshot.text\}\par
\smallskip
\noindent CURRENT URL: \{url\}\par
\smallskip
\noindent DECISION CHECKPOINT: If all active instructions are satisfied now, put send\_msg\_to\_user('done: ...') inside an ```action fence.\par
\smallskip
\noindent LAST ERROR: \{last\_error\}\par
\smallskip
\noindent PAGE CONTENT:\par
\noindent \{axtree\_short\}\par
}\medskip

\end{prompt}

\subsubsection{Milestone Triggering}
\begin{prompt}
\raggedright
\sloppy
\setlength{\emergencystretch}{3em}


\textbf{System}\par\smallskip
{\ttfamily
\noindent You are a milestone trigger judge for a dynamic GUI benchmark.\par
\noindent Your job: decide whether the current page indicates that a specific semantic stage has been reached.\par
\smallskip
\noindent RULES:\par
\noindent - This is a factual judgment about the current page state.\par
\noindent - Look for concrete evidence: listings, filtered results, specific content, navigation state, etc.\par
\noindent - The semantic stage describes a CHECKPOINT in the user's task progress, not the exact UI method. For example, 'search results for sofas are displayed' is satisfied by ANY page showing a list of sofas \textemdash{} whether reached via search, category navigation, or direct URL. Focus on whether the current page state matches the described stage, not on how the agent got there.\par
\noindent - The page must clearly show that the described stage has been achieved.\par
\noindent - When genuinely uncertain, return should\_fire=false. But do not be overly conservative \textemdash{} if the page clearly reaches the stage described by the milestone, fire even if minor details differ from the literal wording.\par
\noindent - Existential vs exclusive: when the milestone describes a result state like 'a list of X is displayed' or 'results showing X', treat it as satisfied if the page contains X candidates the user could act on, EVEN IF the list also contains non-X items. Do not require the list to be exclusively X unless the milestone explicitly says 'only X' / 'exclusively X' / 'purely X'.\par
\noindent - Observable count vs explicit constraint marker: when the milestone describes a quantitative state like 'exactly N stops' or 'N items meeting criteria', you do NOT need the page to literally display 'limited to N' or 'filter applied \ensuremath{\rightarrow} N'. The state is satisfied if N matching items are observably present in the page content. The user does not need the page to advertise its own constraint application.\par
\noindent - Junction vs terminal cleanliness: the milestone marks an intermediate junction on the task path, not the terminal completion of prior instructions. Fire as soon as the junction's observable evidence appears; do not wait for the prior task to be 'fully consummated' or for the page to settle into a single canonical end-state.\par
\noindent - confidence below 0.8 \ensuremath{\rightarrow} should\_fire must be false.\par
\smallskip
}\medskip


\textbf{User}\par\smallskip
{\ttfamily
\noindent Semantic stage to detect:\par
\noindent \{milestone.semantic\_concept\}\par
\smallskip
\noindent Current page content:\par
\noindent \{axtree\_short\}\par
\smallskip
\noindent Has the current page reached the described semantic stage?\par
\noindent List the specific evidence you found (or did not find), then decide.\par
\noindent Return JSON only: \{"should\_fire": bool, "confidence": 0.0-1.0, "reason": "..."\}\par
}\medskip

\end{prompt}

\subsubsection{Instruction Rewriting}
\begin{prompt}
\raggedright
\sloppy
\setlength{\emergencystretch}{3em}


\textbf{System}\par\smallskip
{\ttfamily
\noindent You rewrite dynamic GUI task instructions faithfully.\par
}\medskip


\textbf{User}\par\smallskip
{\ttfamily
\noindent Rewrite the current multi-turn GUI task into ONE complete, user-facing task instruction.\par
\smallskip
\noindent Rules:\par
\noindent - Preserve every active requirement.\par
\noindent - Do not add new requirements.\par
\noindent - For [CHANGED] updates, replace or relax only the conflicting earlier requirement; keep unrelated requirements.\par
\noindent - For [CONDITIONAL] updates, keep the condition in the rewritten instruction. Do not decide the branch during rewriting.\par
\noindent - Do not mention revisions, turns, messages, or this rewrite step.\par
\noindent - Output only the rewritten instruction, no bullets, no explanation.\par
\smallskip
\noindent Current active instructions:\par
\noindent \{structured\}\par
}\medskip

\end{prompt}

\subsection{Evaluation}

\subsubsection{Claim Generation}
\begin{prompt}
\raggedright
\sloppy
\setlength{\emergencystretch}{3em}


\textbf{System}\par\smallskip
{\ttfamily
\noindent You decompose a single atomic task constraint into a list of binary sub-claims. The constraint is satisfied IFF all sub-claims are satisfied (logical AND). Each sub-claim must be independently checkable.\par
\smallskip
\noindent INPUTS you will receive:\par
\noindent - The complete instruction timeline: I0 and revisions in order\par
\noindent - A constraint description (one atomic requirement)\par
\noindent - The agent's full final answer (done\_message)\par
\smallskip
\noindent Interpret the current constraint in the context of the instruction timeline. Preserve its scope and applicable revisions; generate sub-claims only for the current constraint, without adding requirements. Later overrides replace earlier requirements on the same dimension; do not reintroduce superseded requirements or combine mutually exclusive conditional alternatives. Use the timeline as context, not as a checklist to add unrelated constraints.\par
\smallskip
\noindent Each sub-claim has a `type`:\par
\smallskip
\noindent - type='simple': resolvable by reading done\_message text and applying logic. No page evidence needed. Examples of simple predicates:\par
\noindent \hspace*{2.0em}'done\_message identifies exactly 3 distinct items'\par
\noindent \hspace*{2.0em}'the 3 series labels in done\_message are pairwise distinct'\par
\noindent \hspace*{2.0em}'Injustice: Gods Among Us Year One does not appear in the items '\par
\noindent \hspace*{3.0em}'listed in done\_message'\par
\noindent \hspace*{2.0em}'done\_message provides exactly 3 separate URLs, one per item'\par
\smallskip
\noindent - type='page': requires actual page content from the agent's trajectory. Provide subject + optional assertion. The claim is that the subject has the stated property; evidence need not come from the subject's detail page. Examples of page predicates:\par
\noindent \hspace*{2.0em}subject='DCeased', assertion='heavily features Batman/Bat-family'\par
\noindent \hspace*{3.0em}\ensuremath{\rightarrow} evidence that DCeased has this property.\par
\noindent \hspace*{2.0em}subject='Batman: The Dark Knight', assertion='by Tom Taylor'\par
\noindent \hspace*{3.0em}\ensuremath{\rightarrow} evidence that it is by Tom Taylor.\par
\noindent \hspace*{2.0em}subject='Injustice', assertion='page is the homepage of Injustice on       DC.com'\par
\noindent \hspace*{2.0em}subject='sephora.com', assertion=null\par
\noindent \hspace*{3.0em}\ensuremath{\rightarrow} trajectory must show a page on sephora.com (no further check).\par
\smallskip
\noindent DECOMPOSITION RULES:\par
\noindent 1. EXHAUSTIVE: cover every verifiable dimension of the constraint. Missing a dimension means false positives slip through.\par
\noindent 2. DISJOINT: don't repeat the same check.\par
\noindent 3. PREFER SIMPLE: when the same check can be done from done\_message text alone OR via a page, prefer simple. Use page only when text is genuinely insufficient (per-item factual properties, byline, synopsis content, URL semantics, trajectory presence).\par
\noindent 4. PER-ITEM \ensuremath{\times} LIST decomposition: for constraints of the form 'each X must Y' or 'find N X with property Y':\par
\noindent \hspace*{2.0em}- Emit one PAGE sub-claim per item the agent listed (subject + assertion describing Y).\par
\noindent \hspace*{2.0em}- Emit any SIMPLE sub-claims about the list as a whole (count, pairwise distinctness, exclusion).\par
\noindent 5. ANSWER-SECTION SCOPING: done\_message is often a structured document. The agent may write multiple sections \textemdash{} one for each constraint or sub-question of the task. When a constraint says 'each X', 'N X', 'exclude Y', or otherwise refers to a list of items, IDENTIFY THE SPECIFIC SECTION the agent designated as the answer for THIS constraint, and scope your decomposition to that section only. Do NOT include items from other sections (reference items, items mentioned for context, items answering a different constraint) as if they were part of the answer.\par
\noindent \hspace*{2.0em}Section signals to look for in done\_message:\par
\noindent \hspace*{3.0em}- Bold headers like '**Three Batman comic books ...:**' marking what the agent considers the answer to a particular constraint.\par
\noindent \hspace*{3.0em}- Numbered lists '1. ... 2. ... 3. ...' \textemdash{} typically the answer items.\par
\noindent \hspace*{3.0em}- Explicit exclusion phrases like '(excluding X)' inside a header \textemdash{} these CONFIRM the agent has already excluded X from this section's items.\par
\noindent \hspace*{3.0em}- Standalone reference URLs / context paragraphs separated from the numbered list \textemdash{} usually NOT part of the answer count.\par
\noindent \hspace*{2.0em}Example (from a real task \textemdash{} see worked examples below):\par
\noindent \hspace*{3.0em}done\_message has:\par
\noindent \hspace*{4.0em}'**Injustice: Gods Among Us Year One** (the prequel ...):\par
\noindent \hspace*{4.5em}https://www.dc.com/...year-one...\par
\noindent \hspace*{4.5em}**Three Batman comic books by Tom Taylor (different series,          excluding Injustice):**\par
\noindent \hspace*{4.5em}1. Detective Comics \#1090 ...\par
\noindent \hspace*{4.5em}2. Batman \textendash{} One Bad Day ...\par
\noindent \hspace*{4.5em}3. Batman: The Dark Knight ...'\par
\noindent \hspace*{3.0em}For the constraint 'Find exactly three comic books', the answer items are the THREE numbered comics \textemdash{} NOT four. Injustice appears in a separate section answering a different constraint ('provide the link to Year One on DC.com'); it is reference, not part of the count.\par
\noindent \hspace*{3.0em}For the constraint 'Each comic must feature Batman', the 'each' refers to the THREE in the numbered list \textemdash{} NOT to Injustice. Decompose 3 page sub-claims (one per numbered comic), NOT 4.\par
\noindent \hspace*{3.0em}For the constraint 'Exclude Injustice itself', the 3 numbered comics are the candidate set; agent's header explicitly says 'excluding Injustice'; emit one simple sub-claim 'none of the 3 numbered Batman comics is Injustice: Gods Among Us Year One' \textemdash{} do NOT treat Injustice's own reference link as 'inclusion'.\par
\noindent 6. AGENT SAID NOTHING: if the agent did not address this constraint at all in done\_message, decomposition still proceeds \textemdash{} emit the sub-claims the constraint would require regardless. simple sub-claims checking 'done\_message says X' will fail naturally; page sub-claims for process-style requirements may still resolve from trajectory.\par
\smallskip
\noindent FIELDS for type='page':\par
\smallskip
\noindent subject: A VERBATIM SUBSTRING of done\_message that identifies the item. The subject must appear EXACTLY AS IT APPEARS in done\_message \textemdash{} do NOT reformat, do NOT strip punctuation (\#, \textendash{}, :, etc.), do NOT lowercase, do NOT hyphenate, do NOT pull a path segment out of a URL if a readable name is also available.\par
\smallskip
\noindent Self-check before writing subject: can you find the subject as a contiguous substring of done\_message? If no, you reformulated \textemdash{} pick a real verbatim substring instead.\par
\smallskip
\noindent Preference among multiple valid verbatim substrings:\par
\noindent \hspace*{1.0em}1. When agent provides BOTH a human-readable name AND a URL for an item, USE THE NAME. Names appear as display text in AXTrees; URL slugs do not.\par
\noindent \hspace*{1.0em}2. Pick the LONGEST distinctive verbatim substring specific to the item. Avoid generic series names that match many pages.\par
\noindent \hspace*{1.0em}3. URL substrings are only acceptable when no readable name is available (e.g. agent supplied only a link). Even then, prefer a short distinctive verbatim token over the full URL.\par
\smallskip
\noindent Examples:\par
\noindent \hspace*{1.0em}done\_message: '1. **Detective Comics \#1090** (ongoing series) - https://www.dc.com/comics/detective-comics-2016/detective-comics-1090'\par
\noindent \hspace*{2.0em}\ensuremath{\checkmark} subject='Detective Comics \#1090'  (verbatim, with \#, name preferred)\par
\noindent \hspace*{2.0em}\ensuremath{\times} subject='Detective Comics 1090'   (NOT verbatim \textemdash{} \# was dropped)\par
\noindent \hspace*{2.0em}\ensuremath{\times} subject='Detective Comics'        (verbatim BUT generic; matches many DC pages)\par
\noindent \hspace*{2.0em}\ensuremath{\times} subject='detective-comics-1090'   (URL slug \textemdash{} verbatim but doesn't grep-hit display text; name is available, prefer it)\par
\smallskip
\noindent \hspace*{1.0em}done\_message: 'Batman: The Dark Knight (2021 miniseries)'\par
\noindent \hspace*{2.0em}\ensuremath{\checkmark} subject='Batman: The Dark Knight'   (verbatim substring)\par
\noindent \hspace*{2.0em}\ensuremath{\times} subject='Batman: The Dark Knight (2021 miniseries)'   (verbatim BUT includes parenthetical that varies across pages)\par
\noindent \hspace*{2.0em}\ensuremath{\times} subject='Batman: The Dark Knight 2021 miniseries'   (NOT verbatim \textemdash{} parenthesis changed)\par
\smallskip
\noindent \hspace*{1.0em}done\_message: 'https://www.dc.com/search?q=Injustice\%3A+Year+One' (no readable name attached)\par
\noindent \hspace*{2.0em}\ensuremath{\checkmark} subject='Injustice\%3A+Year+One' or 'Injustice'   (verbatim URL substrings; agent gave no name)\par
\noindent \hspace*{2.0em}\ensuremath{\times} subject='Injustice: Year One'   (NOT verbatim \textemdash{} URL-decoded)\par
\smallskip
\noindent assertion: the property the subject must have. THREE cases:\par
\noindent \hspace*{1.0em}(a) Agent wrote a specific say-so verbatim: COPY EXACTLY.\par
\noindent \hspace*{3.0em}done\_message: 'DCeased (heavily features Batman/Bat-family)'\par
\noindent \hspace*{3.0em}\ensuremath{\rightarrow} assertion='heavily features Batman/Bat-family'\par
\noindent \hspace*{1.0em}(b) Constraint requires a per-item attribute that agent claimed only globally (or did not claim per item, but the attribute is verifiable as a property of each item): use a SHORT canonical phrase for that property.\par
\noindent \hspace*{3.0em}Constraint: 'Each comic by Tom Taylor', agent: 'Author: Tom Taylor (one global statement)'\par
\noindent \hspace*{3.0em}\ensuremath{\rightarrow} For each item: assertion='by Tom Taylor' (canonical, not verbatim agent text \textemdash{} but only used because the per-item byline is the property that must be demonstrated).\par
\noindent \hspace*{1.0em}(c) Subject just needs to exist / be visited (count, type, trajectory-presence constraints): assertion=null.\par
\noindent \hspace*{3.0em}Constraint: 'Find 3 comic books' \ensuremath{\rightarrow} for each subject, assertion=null (the page just needs to show it's a real comic).\par
\noindent \hspace*{3.0em}Constraint: 'on sephora.com' \ensuremath{\rightarrow} subject='sephora.com', assertion=null.\par
\smallskip
\noindent DO NOT FABRICATE: do not synthesize per-item assertions from the constraint wording when the constraint is about the LIST AS A WHOLE (not per-item).\par
\noindent \hspace*{1.0em}Constraint: 'Find exactly three comic books'\par
\noindent \hspace*{1.0em}WRONG: per-item assertion='one of three comic books' \textemdash{} fabricated; the COUNT is a list-level property, handled by a simple sub-claim.\par
\noindent \hspace*{1.0em}RIGHT: emit one simple sub-claim 'done\_message lists exactly 3 items' + one page sub-claim per item with assertion=null.\par
\smallskip
\noindent Worked examples:\par
\smallskip
\noindent Constraint: 'Find exactly three comic books'\par
\noindent done\_message lists Batman: The Dark Knight, Batman: The Detective, DCeased\par
\noindent Sub-claims:\par
\noindent \hspace*{1.0em}\{type: simple, description: 'done\_message identifies exactly 3 distinct comic books'\}\par
\noindent \hspace*{1.0em}\{type: page, subject: 'Batman: The Dark Knight', assertion: null, description: 'Batman: The Dark Knight is shown to be a real comic '\par
\noindent \hspace*{2.0em}'book on the trajectory'\}\par
\noindent \hspace*{1.0em}\{type: page, subject: 'Batman: The Detective', assertion: null, description: 'Batman: The Detective is shown to be a real comic '\par
\noindent \hspace*{2.0em}'book on the trajectory'\}\par
\noindent \hspace*{1.0em}\{type: page, subject: 'DCeased', assertion: null, description: 'DCeased is shown to be a real comic book on the trajectory'\}\par
\smallskip
\noindent Constraint: 'Each comic must feature Batman'\par
\noindent done\_message: '... 1. Batman: The Dark Knight (2021) 2. Batman: The Detective (2022) 3. DCeased (heavily features Batman/Bat-family)'\par
\noindent Sub-claims:\par
\noindent \hspace*{1.0em}\{type: page, subject: 'Batman: The Dark Knight', assertion: null, description: 'Batman: The Dark Knight features Batman (title self-evident)'\}\par
\noindent \hspace*{1.0em}\{type: page, subject: 'Batman: The Detective', assertion: null, description: 'Batman: The Detective features Batman (title self-evident)'\}\par
\noindent \hspace*{1.0em}\{type: page, subject: 'DCeased', assertion: 'heavily features Batman/Bat-family', description: 'DCeased page corroborates the agent statement'\}\par
\smallskip
\noindent Constraint: 'Each comic from a different series'\par
\noindent done\_message: '1. Batman: The Dark Knight (2021 miniseries), 2. Batman: The Detective (2022 miniseries), 3. DCeased (2019-2021 series)'\par
\noindent Sub-claims:\par
\noindent \hspace*{1.0em}\{type: simple, description: 'the 3 series labels in done\_message '\par
\noindent \hspace*{2.0em}'(2021 miniseries / 2022 miniseries / 2019-2021 series) are '\par
\noindent \hspace*{2.0em}'pairwise distinct'\}\par
\noindent \hspace*{1.0em}\{type: page, subject: 'Batman: The Dark Knight', assertion: '2021 miniseries', description: 'Dark Knight page corroborates 2021 miniseries series label'\}\par
\noindent \hspace*{1.0em}\{type: page, subject: 'Batman: The Detective', assertion: '2022 miniseries', description: 'Detective page corroborates 2022 miniseries label'\}\par
\noindent \hspace*{1.0em}\{type: page, subject: 'DCeased', assertion: '2019-2021 series', description: 'DCeased page corroborates 2019-2021 series label'\}\par
\smallskip
\noindent Constraint: 'Exclude Injustice: Gods Among Us Year One itself'\par
\noindent Sub-claims:\par
\noindent \hspace*{1.0em}\{type: simple, description: 'Injustice: Gods Among Us Year One does not appear among the items listed in done\_message'\}\par
\smallskip
\noindent Constraint: 'Provide a link to the homepage of Year One on DC.com'\par
\noindent done\_message: 'Links: https://www.dc.com/search?q=Injustice\%3A+Year+One'\par
\noindent Sub-claims:\par
\noindent \hspace*{1.0em}\{type: page, subject: 'search?q=Injustice', assertion: 'page is the homepage of Injustice: Gods Among Us Year One on DC.com', description: 'the URL agent provided resolves to the comic homepage on DC.com'\}\par
\smallskip
\noindent Constraint: 'Provide a link to the homepage of each comic on DC.com'\par
\noindent done\_message: 'Links: https://www.dc.com/search?q=Injustice...' (only one link given)\par
\noindent Sub-claims:\par
\noindent \hspace*{1.0em}\{type: simple, description: 'done\_message provides 3 distinct homepage URLs, one per comic book'\}\par
\noindent \hspace*{1.0em}(no page sub-claims \textemdash{} agent did not provide per-comic links; the simple sub-claim already fails)\par
\smallskip
\noindent Output JSON only:\par
\noindent \{"sub\_claims": [\par
\noindent \hspace*{1.0em}\{"claim\_id": 1, "type": "simple", "description": "..."\},\par
\noindent \hspace*{1.0em}\{"claim\_id": 2, "type": "page", "description": "...", "subject": "...", "assertion": "..." | null\}\par
\noindent ]\}\par
}\medskip


\textbf{User}\par\smallskip
{\ttfamily
\noindent Complete instruction timeline:\par
\noindent \{instruction\_timeline\}\par
\smallskip
\noindent Constraint:\par
\noindent \{constraint\_description\}\par
\noindent Task was executed on: \{execution\_date\}\par
\smallskip
\noindent done\_message:\par
\noindent \{done\_message\}\par
\smallskip
\noindent Decompose the constraint into binary sub-claims.\par
}\medskip

\end{prompt}

\subsubsection{Simple Claim Verification}
\begin{prompt}
\raggedright
\sloppy
\setlength{\emergencystretch}{3em}


\textbf{System}\par\smallskip
{\ttfamily
\noindent You verify whether a given binary predicate is correct, given the agent's full final answer (done\_message) and the original task constraint as context. The predicate is checkable by reading done\_message and applying logic or simple factual reasoning. You do NOT have access to web pages \textemdash{} your judgment must come from the text you are given.\par
\smallskip
\noindent GENERAL RULES:\par
\noindent 1. Examine the predicate carefully. Decide satisfied=true only when done\_message demonstrably supports it; satisfied=false when it is contradicted or absent.\par
\noindent 2. The original task constraint is provided as context \textemdash{} it tells you why the predicate matters, but the predicate itself is what you must judge.\par
\noindent 3. Reasoning must cite specific text from done\_message. Quote snippets using single quotes when relevant.\par
\noindent 4. Be strict about format predicates: 'lists exactly 3' means 3, not 2 or 4. 'Provides 3 distinct URLs' means 3 separate URLs, not one URL repeated.\par
\noindent 5. Apply fuzzy matching for minor variations (case, spacing, hyphenation, common abbreviations) \textemdash{} but do not stretch matches across genuinely different items.\par
\noindent 6. If done\_message does not contain the information needed to judge the predicate, return satisfied=false (absence of evidence for an assertion is failure, not insufficiency).\par
\noindent 7. ANSWER-SECTION SCOPING. done\_message is often a structured document with multiple sections \textemdash{} one per constraint or sub-question. When the predicate involves COUNT ('lists exactly N'), EXCLUSION ('X does not appear in the items'), DISTINCTNESS, or any other property OVER A LIST OF ITEMS, you MUST first identify which section of done\_message the agent designated as the answer for THIS constraint. Apply the predicate ONLY to items in that section.\par
\noindent \hspace*{2.0em}Section signals to look for:\par
\noindent \hspace*{3.0em}- Bold headers like '**Three Batman comic books ...:**' marking what the agent considers the answer to a particular constraint.\par
\noindent \hspace*{3.0em}- Numbered lists '1. ... 2. ... 3. ...' \textemdash{} typically the answer items.\par
\noindent \hspace*{3.0em}- Explicit exclusion phrases like '(excluding X)' inside a header \textemdash{} these CONFIRM the agent has already excluded X from this section's items.\par
\noindent \hspace*{3.0em}- Standalone reference URLs / context paragraphs separated from the numbered list \textemdash{} usually NOT part of the answer count.\par
\noindent \hspace*{2.0em}Worked example. Suppose done\_message has:\par
\noindent \hspace*{4.0em}'**Injustice: Gods Among Us Year One** (the prequel ...):\par
\noindent \hspace*{4.5em}https://www.dc.com/...year-one...\par
\noindent \hspace*{4.5em}**Three Batman comic books by Tom Taylor (different          series, excluding Injustice):**\par
\noindent \hspace*{4.5em}1. Detective Comics \#1090 ...\par
\noindent \hspace*{4.5em}2. Batman \textendash{} One Bad Day ...\par
\noindent \hspace*{4.5em}3. Batman: The Dark Knight ...'\par
\noindent \hspace*{3.0em}Predicate 'done\_message identifies exactly 3 distinct comic books': the answer section is the numbered list (3 items) \ensuremath{\rightarrow} satisfied=true. DO NOT count Injustice as a 4th item \textemdash{} it appears in a SEPARATE section that answers a different constraint.\par
\noindent \hspace*{3.0em}Predicate 'Injustice: Gods Among Us Year One does not appear among the items listed': the relevant 'items' are the numbered list (the agent's answer to the find-three constraint); Injustice is NOT in that numbered list \ensuremath{\rightarrow} satisfied=true. The agent's separate reference to Injustice (with its own URL, answering the link constraint) does NOT mean it appears 'among the items' for this exclusion predicate.\par
\noindent \hspace*{3.0em}Predicate 'done\_message provides exactly 3 distinct homepage URLs, one per comic book': count URLs in the THREE numbered comics' section, not URLs from the Injustice reference section.\par
\smallskip
\noindent Output JSON only: \{"satisfied": bool, "reason": "...", "confidence": 0.0-1.0\}\par
}\medskip


\textbf{User}\par\smallskip
{\ttfamily
\noindent Original constraint (context):\par
\noindent \{constraint\_description\}\par
\smallskip
\noindent Task was executed on: \{execution\_date\}\par
\smallskip
\noindent Predicate to verify:\par
\noindent \{predicate\}\par
\smallskip
\noindent Agent's done\_message:\par
\noindent \{done\_message\}\par
\smallskip
\noindent Is the predicate satisfied? Cite specific text from done\_message.\par
}\medskip

\end{prompt}

\subsubsection{Conditional Branch Resolution}
\begin{prompt}
\raggedright
\sloppy
\setlength{\emergencystretch}{3em}


\textbf{System}\par\smallskip
{\ttfamily
\noindent You judge whether an agent's done\_message addresses a specific page-observable condition. The condition came from a conditional override instruction (e.g. 'if X, then Y') \textemdash{} the agent was expected to check it on the page and report observation in the done\_message.\par
\smallskip
\noindent Return one of three states:\par
\noindent \hspace*{1.0em}'true'          \textemdash{} done\_message clearly states the condition HOLDS, with concrete values from the page.\par
\noindent \hspace*{1.0em}'false'         \textemdash{} done\_message clearly states the condition DOES NOT HOLD, with concrete values from the page.\par
\noindent \hspace*{1.0em}'not\_addressed' \textemdash{} done\_message does not address the condition. Default to this when uncertain \textemdash{} do NOT guess.\par
\smallskip
\noindent RULES:\par
\noindent 1. Look for explicit observation statements. Phrases like '[CONDITIONAL OBSERVATION]', 'I observed', 'the price difference is X', 'I compared X and Y' are strong signals the agent addressed it.\par
\noindent 2. The agent's wording does not need to match the condition verbatim. 'price difference is 160 RMB, below threshold' addresses 'price difference exceeds 500 RMB' \ensuremath{\rightarrow} state='false'. '900 RMB diff exceeds 500' \ensuremath{\rightarrow} state='true'.\par
\noindent 3. Concrete values are required. 'I saw the prices' without numbers or comparison is not\_addressed. Vague claims about thresholds without actual values are not\_addressed.\par
\noindent 4. The agent's selection is NOT used here \textemdash{} only what the agent SAID the condition state was. Inconsistencies between what the agent reports and what the agent did get caught by downstream sub\_claim judgment, not here.\par
\noindent 5. Reasoning must cite specific text from done\_message.\par
\smallskip
\noindent Output JSON only: \{"condition\_state": "true"|"false"|"not\_addressed", "reason": "...", "confidence": 0.0-1.0\}\par
}\medskip


\textbf{User}\par\smallskip
{\ttfamily
\noindent Condition to evaluate:\par
\noindent \{condition\_text\}\par
\smallskip
\noindent Task was executed on: \{execution\_date\}\par
\smallskip
\noindent Agent's done\_message:\par
\noindent \{done\_message\}\par
\smallskip
\noindent Did the agent address the condition? If yes, did the agent report it as true or false?\par
}\medskip

\end{prompt}

\subsubsection{Web Evidence Verification}
\begin{prompt}
\raggedright
\sloppy
\setlength{\emergencystretch}{3em}


\textbf{System}\par\smallskip
{\ttfamily
\noindent You verify a page sub-claim against a GUI page snapshot (AXTree), in the context of a task constraint. Your judgment must come from the AXTree and the supplied page URL \textemdash{} do NOT use world knowledge or background facts to fill gaps. The URL may help identify the page; it does not by itself prove that an activity or property is present.\par
\smallskip
\noindent A sub-claim has two parts:\par
\noindent \hspace*{1.0em}subject \textemdash{} the item the page should be about (book, product, person, URL fragment, domain, etc.)\par
\noindent \hspace*{1.0em}agent\_assertion \textemdash{} what the page must show about the subject. May be null when the sub-claim only requires the subject to be shown, with no further attribute check.\par
\smallskip
\noindent Decide three things:\par
\smallskip
\noindent 1. is\_subject\_page: Can this frame host a claim about the subject? True when the page is the subject's detail / content / profile / home page, or is otherwise centered on the subject (e.g. for domain subjects like 'sephora.com', any page on that host counts). Also true when this is a listing / search / filter / results view and the subject appears as a result item (card, row, or hit). False when the subject only appears as a navigation menu item, an unrelated link, or a passing mention.\par
\smallskip
\noindent 2. assertion\_supported: If agent\_assertion is provided, does THIS SAME frame show that the subject has that property? True when the page contains direct textual evidence (synopsis text, attribute labels, character lists, byline, price tags, series labels), OR when the subject is a result item here and the view's visible filters / query / category / chips entail the property. False when this frame does not contain such evidence \textemdash{} even if the assertion happens to be true in the world. Null when agent\_assertion itself is null.\par
\smallskip
\noindent 3. evidence\_summary: One or two short sentences describing what this page shows about the subject relevant to the constraint. Quote or paraphrase short snippets directly from the AXTree or page URL. If the page is not about the subject, summarize what the page actually is. This summary is used in diagnostic reasons \textemdash{} it should be specific enough that a reader can reconstruct why you decided as you did.\par
\smallskip
\noindent STRICTNESS:\par
\noindent - Apply fuzzy matching for minor variations (case, spacing, hyphenation, year format, common abbreviations) \textemdash{} but final decisions must rely on explicit AXTree evidence and page URL context.\par
\noindent - If the page is entirely irrelevant or inaccessible (error page, blank, blocked), conclude is\_subject\_page=false directly.\par
\noindent - For null agent\_assertion: is\_subject\_page=true is sufficient \textemdash{} you do not need to verify any further attribute.\par
\smallskip
\noindent Examples:\par
\noindent - subject='Batman: The Dark Knight', agent\_assertion='2021 miniseries by Tom Taylor'\par
\noindent \hspace*{1.0em}AXTree shows a comic detail page with the title, '2021 miniseries' label, and 'by Tom Taylor' byline \ensuremath{\rightarrow}\par
\noindent \hspace*{2.0em}is\_subject\_page=true, assertion\_supported=true, evidence\_summary='detail page identifies the comic as Batman: The Dark Knight (2021 miniseries) by Tom Taylor'.\par
\noindent - subject='DCeased', agent\_assertion='heavily features Batman/Bat-family'\par
\noindent \hspace*{1.0em}AXTree shows the DCeased detail page with a synopsis about a zombie outbreak, no mention of Batman characters \ensuremath{\rightarrow}\par
\noindent \hspace*{2.0em}is\_subject\_page=true, assertion\_supported=false, evidence\_summary='detail page identifies DCeased (2019-2021 ongoing series); synopsis describes a zombie outbreak with no Batman/Bat-family characters visible'.\par
\noindent - subject='American Hairless Terrier', agent\_assertion='medium barking'\par
\noindent \hspace*{1.0em}AXTree is a breed listing with Barking Level 'Medium' checked and a result card 'American Hairless Terrier' \ensuremath{\rightarrow}\par
\noindent \hspace*{2.0em}is\_subject\_page=true, assertion\_supported=true, evidence\_summary='breed is a result item on a listing filtered to medium barking'.\par
\noindent - subject='sephora.com', agent\_assertion=null\par
\noindent \hspace*{1.0em}AXTree shows a page with URL containing sephora.com \ensuremath{\rightarrow}\par
\noindent \hspace*{2.0em}is\_subject\_page=true, assertion\_supported=null, evidence\_summary='page is on sephora.com (URL visible in nav)'.\par
\noindent - subject='search?q=Injustice', agent\_assertion='homepage of Injustice on DC.com'\par
\noindent \hspace*{1.0em}AXTree shows a search results listing on dc.com \ensuremath{\rightarrow}\par
\noindent \hspace*{2.0em}is\_subject\_page=true, assertion\_supported=false, evidence\_summary='page is a dc.com search results listing for Injustice, not a comic homepage'.\par
\smallskip
\noindent Respond with JSON only: \{"is\_subject\_page": bool, "assertion\_supported": bool | null, "evidence\_summary": "...", "confidence": 0.0-1.0, "reason": "..."\}\par
}\medskip


\textbf{User}\par\smallskip
{\ttfamily
\noindent Constraint:\par
\noindent \{constraint\_description\}\par
\smallskip
\noindent Sub-claim to verify:\par
\noindent \hspace*{1.0em}subject: \{subject\}\par
\noindent \hspace*{1.0em}\{assertion\_line\}\par
\noindent Task was executed on: \{execution\_date\}\par
\smallskip
\noindent Page URL (full):\par
\noindent \{page\_url\}\par
\smallskip
\noindent Page state (AXTree):\par
\noindent \{\_truncate\_axtree(axtree)\}\par
\smallskip
\noindent Verify the sub-claim against the AXTree and page URL.\par
}\medskip


\textbf{User: \{assertion\_line\}, with assertion}\par\smallskip
{\ttfamily
\noindent agent\_assertion: \{agent\_assertion\}\par
\smallskip
}\medskip


\textbf{User: \{assertion\_line\}, without assertion}\par\smallskip
{\ttfamily
\noindent agent\_assertion: null (only the subject's presence on the page needs to be verified)\par
\smallskip
}\medskip

\end{prompt}

\subsubsection{Post-Revision Behavioral Assessment (WRA)}
\begin{prompt}
\raggedright
\sloppy
\setlength{\emergencystretch}{3em}


\textbf{System}\par\smallskip
{\ttfamily
\noindent You judge whether a GUI agent's actions in a short window (1-3 steps) responded to a specific user instruction that was just injected. The instruction is one of a sequence of corrections the user gave the agent during execution.\par
\smallskip
\noindent Decide TWO things, independently:\par
\smallskip
\noindent 1. wra \ensuremath{\in} \{0, 1\}\par
\noindent \hspace*{1.5em}wra=1 \textemdash{} at least one action in the window concretely advances the new instruction's intent (issuing a relevant search, navigating to a relevant page, filtering / refining results to match the new constraint, opening a candidate that satisfies the new requirement, etc.).\par
\noindent \hspace*{1.5em}wra=0 \textemdash{} none of the actions clearly serve the new instruction. Continuing prior unrelated work, looping on the same failed action, idling, or making changes orthogonal to the instruction all count as wra=0.\par
\noindent \hspace*{1.5em}Be lenient on partial / exploratory progress (clicking a filter that partly addresses the new requirement is wra=1) but strict on irrelevance (any-website search that ignores the instruction's specifics is wra=0).\par
\smallskip
\noindent 2. adaptation \ensuremath{\in} \{overwrite, additive, unrelated\}\par
\noindent \hspace*{1.5em}overwrite \textemdash{} the agent abandons the page / result it was on before injection and re-navigates from scratch (back to a search engine, a different site, or a fresh query). Characterized by goto() to a new domain, opening a new search, or discarding the current candidate.\par
\noindent \hspace*{1.5em}additive \textemdash{} the agent extends or refines the work in place: applies a filter, scrolls, opens a sub-link of the current result, edits the existing query, etc. The previous progress is preserved.\par
\noindent \hspace*{1.5em}unrelated \textemdash{} actions don't engage with the instruction at all (e.g. re-issuing a done call, looping on the same prior click, navigating to something orthogonal). Use this label freely when wra=0 and the actions don't fit overwrite/additive.\par
\smallskip
\noindent RULES:\par
\noindent - Judge only on the actions / URLs given. Do not assume the agent will do something later.\par
\noindent - The instruction text is verbatim from the user. Read it carefully \textemdash{} 'exactly three' means the agent should be acting on the count requirement, not anything else in the prompt.\par
\noindent - A done call inside the window is not a response by itself unless the prior actions already satisfied the instruction.\par
\noindent - Reasoning must reference specific step actions / URLs.\par
\smallskip
\noindent Output JSON only: \{"wra": 0|1, "adaptation": "overwrite"|"additive"|"unrelated", "reason": "...", "confidence": 0.0-1.0\}\par
}\medskip


\textbf{User}\par\smallskip
{\ttfamily
\noindent Injected instruction (\{instruction\_id\}):\par
\noindent \{instruction\_text\}\par
\smallskip
\noindent Task was executed on: \{execution\_date\}\par
\smallskip
\noindent Agent URL just before injection: \{pre\_injection\_url\}\par
\noindent Action window after injection (chronological):\par
\noindent \{\_format\_window(window\_events)\}\par
\smallskip
\noindent Judge wra and adaptation.\par
}\medskip

\end{prompt}